\documentclass{article} %
\usepackage{iclr2024_conference,times}

\usepackage{amsmath,amsfonts,bm}

\def\1{\bm{1}}

\DeclareMathAlphabet{\mathsfit}{\encodingdefault}{\sfdefault}{m}{sl}
\SetMathAlphabet{\mathsfit}{bold}{\encodingdefault}{\sfdefault}{bx}{n}

\newcommand{\E}{\mathbb{E}}

\newcommand{\R}{\mathbb{R}}

\DeclareMathOperator*{\argmin}{arg\,min}

\usepackage{hyperref}
\usepackage{url}

\usepackage{algorithm}
\usepackage{algpseudocode}
\usepackage[export]{adjustbox}
\usepackage{bm}
\usepackage{tabu}
\usepackage{subcaption}
\usepackage{wrapfig}
\usepackage{graphicx}
\usepackage{multirow}
\usepackage{multirow,makecell}
\usepackage{booktabs,siunitx}
\usepackage{amssymb}
\usepackage{amsthm}
\usepackage{pifont}%
\usepackage{thmtools}
\usepackage{thm-restate}
\usepackage[capitalize,noabbrev]{cleveref} %

\DeclareRobustCommand\onedot{\futurelet\@let@token\@onedot}
\def\onedot{.\ } %
\def\eg{\emph{e.g}\onedot} 
\def\ie{\emph{i.e}\onedot}

\definecolor{dark2green}{rgb}{0.1, 0.65, 0.3}

\declaretheorem[name=Proposition]{prop}

\title{AdaFlood: Adaptive Flood Regularization}

\author{%
Wonho Bae\\
University of British Columbia \& Borealis AI\\
\texttt{whbae@cs.ubc.ca}
\And%
Yi Ren\\
University of British Columbia \& Borealis AI\\
\texttt{renyi.joshua@gmail.com}
\And%
Mohamad Osama Ahmed\\
Borealis AI\\
\texttt{mohamad.o.ahmed@borealisai.com}
\And%
Frederick Tung\\
Borealis AI\\
\texttt{frederick.tung@borealisai.com\hspace{3.5mm}}
\And%
Danica J.\ Sutherland\\
University of British Columbia \& Amii \hspace{10.3mm} \\
\texttt{dsuth@cs.ubc.ca}
\And%
Gabriel L.\ Oliveira\\
Borealis AI\\
\texttt{gabriel.oliveria@borealisai.com}
}

\def\docstatus{preprint}

\ifdefstring{\docstatus}{submission}{
}{}

\makeatletter  %
\ifdefstring{\docstatus}{preprint}{
    \iclrfinalcopy
    \let\mytitle\@title
    \patchcmd{\@maketitle}{\lhead{Published as a conference paper at ICLR 2024}}{\chead{\textsc{\mytitle}}}{}{}
}{}
\makeatother

\ifdefstring{\docstatus}{final}{
    \iclrfinalcopy
}

\begin{document}

\maketitle
\ifdefstring{\docstatus}{preprint}{\thispagestyle{plain}}{}

\begin{abstract}
Although neural networks are conventionally optimized towards zero training loss, it has been recently learned that targeting a non-zero training loss threshold, referred to as a flood level, often enables better test time generalization. 
Current approaches, however, apply the same constant flood level to all training samples, which inherently assumes all the samples have the same difficulty.
We present AdaFlood, a novel flood regularization method that adapts the flood level of each training sample according to the difficulty of the sample. 
Intuitively, since training samples are not equal in difficulty, the target training loss should be conditioned on the instance. 
Experiments on datasets covering four diverse input modalities -- text, images, asynchronous event sequences, and tabular -- demonstrate the versatility of AdaFlood across data domains and noise levels.
\end{abstract}

\section{Introduction}

Preventing overfitting is an important problem of great practical interest in training deep neural networks, which often have the capacity to memorize entire training sets, even ones with incorrect labels \citep{neyshabur2015search,zhang2021understanding}. 
Common strategies to reduce overfitting and improve generalization performance include weight regularization \citep{krogh1991weight,tibshirani1996regression,lq2010liu2010}, dropout \citep{wager2013dropout,srivastava2014dropout,liang2021rdrop}, label smoothing \citep{yuan2020revisiting}, and data augmentation \citep{balestriero2022effects}. 

Although neural networks are conventionally optimized towards zero training loss, it has recently been shown that targeting a non-zero training loss threshold, referred to as a flood level, provides a surprisingly simple yet effective strategy to reduce overfitting \citep{flood2020ishida,iflood2022xie2022}. 
The original Flood regularizer \citep{flood2020ishida} drives the \emph{mean} training loss towards a constant, non-zero flood level, while the state-of-the-art iFlood regularizer \citep{iflood2022xie2022} applies a constant, non-zero flood level to \textit{each} training instance.

Training samples are, however, not uniformly difficult:
some instances have more irreducible uncertainty than others (\ie heteroskedastic noise),
while some instances are simply easier to fit than others.
It may not be beneficial to aggressively drive down the training loss for training samples that are outliers, noisy, or mislabeled. 
We explore this difference in the difficulty of training samples further in \cref{subsec:background}.
To address this issue, we present Adaptive Flooding (AdaFlood), a novel flood regularizer that adapts the flood level of each training sample according to the difficulty of the sample (\cref{subsec:adaflood}).
We present theoretical support for AdaFlood in \cref{subsec:theory}.
  
Like previous flood regularizers, AdaFlood is simple to implement and compatible with any optimizer. 
AdaFlood determines the appropriate flood level for each sample using an auxiliary network that is trained on a subset of training dataset. 
Adaptive flood levels need to be computed for each instance only once, in a pre-processing step prior to training the main network. 
The results of this pre-processing step are not specific to the main network, and so can be shared across multiple hyper-parameter tuning runs.
Furthermore, we propose a significantly more efficient way to train an auxiliary model based on fine-tuning, which saves substantially in memory and computation, especially for overparameterized neural networks (\cref{subsec:ablation}).

Our experiments (\cref{sec:experiments}) demonstrate that AdaFlood generally outperforms previous flood methods on a variety of tasks, including image and text classification, probability density estimation for asynchronous event sequences, and regression for tabular datasets. 
Models trained with AdaFlood are also more robust to noise (\cref{subsec:noisy})
and better-calibrated (\cref{subsec:calibration})
than those trained with other flood regularizers.

\section{Related Work}
\label{sec:related_work}

Regularization techniques have been broadly explored in the machine learning community to improve the generalization ability of neural networks.
Regularizers augment or modify the training objective and are typically compatible with different model architectures, base loss functions, and optimizers. 
They can be used to achieve diverse purposes including reducing overfitting \citep{hanson1988comparing, ioffe2015batch, krogh1991weight, liang2021rdrop, lim2022noisy, srivastava2014dropout, szegedy2016rethinking, verma2019manifold, yuan2020revisiting, zhang2017mixup}, addressing data imbalance \citep{cao2019imbalance, gong2022ranksim}, and compressing models \citep{ding2019regularizing, li2020group, zhuang2020neuron}. 

AdaFlood is a regularization technique for reducing overfitting. 
Commonly adopted techniques for reducing overfitting include weight decay~\citep{hanson1988comparing,krogh1991weight}, dropout~\citep{liang2021rdrop,srivastava2014dropout}, batch normalization~\citep{ioffe2015batch}, label smoothing~\citep{szegedy2016rethinking, yuan2020revisiting}, and data augmentation~\citep{lim2022noisy,verma2019manifold, zhang2017mixup}.
Inspired by work on overparametrization and double descent~\citep{belkin2019reconciling, nakkiran2021deep}, \citet{flood2020ishida} proposed a technique that aims to prevent the training loss from reaching zero by maintaining a small constant value; they termed this ``flooding'' by analogy to keeping the bottom of a container flooded with water. %
\citet{iflood2022xie2022} investigated instability in flooding, as it can lead to different solutions inconsistent in their generalization abilities and predictions for individual data points. 
They proposed an individual flooding loss function, called iFlood, to suppress confidence on over-fit examples while better fitting under-fitted instances. 
In contrast to the original flood regularizer, which encourages the \textit{overall} training loss towards a constant target, iFlood drives \textit{each} training sample's loss towards some constant $b$.

AdaFlood instead uses an auxiliary model trained on a heldout dataset to assign an adaptive flood level to each training sample. 
Using a heldout dataset to condition the training of the primary model is a well-known strategy in machine learning, and is regularly seen in meta-learning \citep{bertinetto2019meta, franceschi2018bilevel}, batch or data selection \citep{fan2018learning, mindermann2022prioritized}, and neural architecture search \citep{liu2019darts, wang2021rethinking}, among other areas.

\section{Adaptive Flooding}
\label{sec:method}

Adaptive Flooding (AdaFlood) is a general regularization method for training neural networks; it can accommodate any typical loss function and optimizer.

\subsection{Problem Statement}
\label{subsec:background}

\paragraph{Background}
Given a labeled training dataset  $\mathcal{D} = \{(\bm{x}_i, y_i)\}_{i=1}^N$,
where $\bm{x}_i \in \mathcal{X}$ are data samples and $y_i \in  \mathcal{Y}$ are labels, we train a neural network $f :  \mathcal{X} \rightarrow \widehat{\mathcal{Y}}$ by minimizing a training loss $\ell : \mathcal{Y} \times \widehat{\mathcal{Y}} \rightarrow \mathbb{R}$.
In supervised learning we usually have $\ell \ge 0$, but in settings such as density estimation it may be negative.
While conventional training procedures attempt to minimize the average training loss,
this can lead to overfitting on training samples. 

The original flood regularizer \citep{flood2020ishida} defines a global flood level for the average training loss, attempting to reduce the ``incentive'' to overfit. 
Denote the average training loss by $\mathcal{L} = \frac{1}{B} \sum_{i=1}^B \ell(y_i, f(x_i))$, where $f(x_i)$ denotes the model prediction and $B$ is the size of a mini-batch. 
Instead of minimizing $\mathcal{L}$, Flood \citep{flood2020ishida} regularizes the training by minimizing 
\begin{equation}
  \mathcal{L}_{\text{Flood}} = |\mathcal{L} - b| + b \, ,
\end{equation}
where the hyperparameter $b$ is a fixed flood level. 
Individual Flood (iFlood)
instead assigns a ``local'' flood level,
trying to avoid instability observed with Flood \citep{iflood2022xie2022}:
\begin{equation}
  \mathcal{L}_{\text{iFlood}} = \frac{1}{B}\sum_{i=1}^{B}\big(|\ell(y_i, f(\bm{x}_i)) - b| + b\big)
.\end{equation}

\begin{figure*}[t!]
    \begin{subfigure}[t]{0.44\textwidth}
        \centering
        \includegraphics[width=0.99\textwidth]{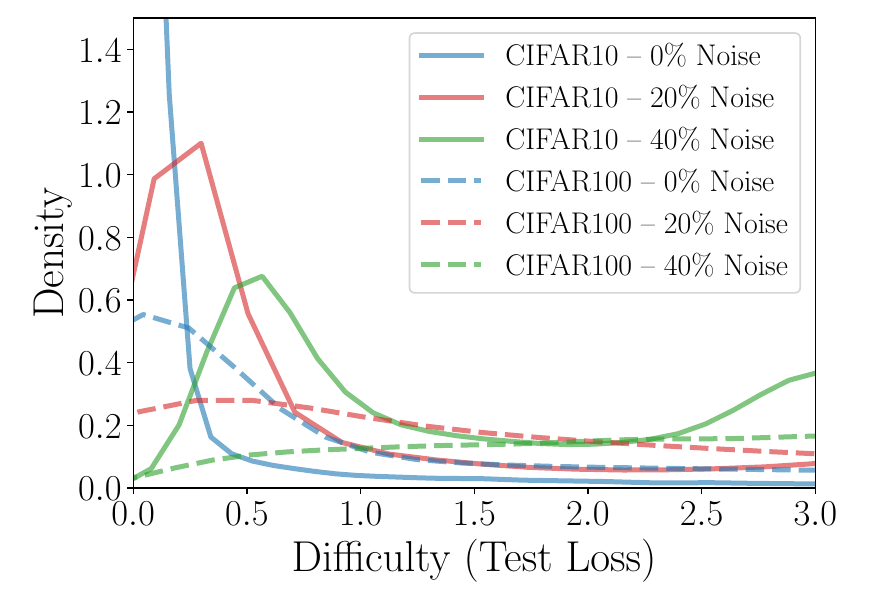}
        \caption{Dispersion of difficulty}
        \label{fig:dispersion_difficulty}
    \end{subfigure}
    \begin{subfigure}[t]{0.55\textwidth}
        \centering
        \includegraphics[width=0.99\textwidth]{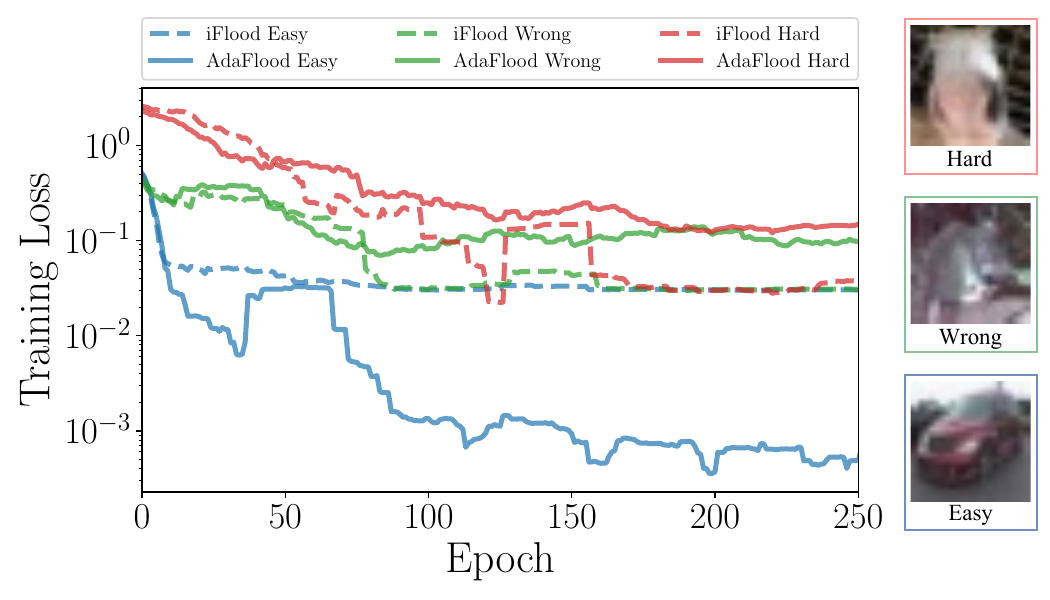}
        \caption{Training dynamics by difficulty}
        \label{fig:varying_difficulty}
    \end{subfigure}
    \caption{(a) Illustration of how difficulties of examples are dispersed with and without label noise (where the relevant portion of examples have their label switched to a random other label). (b) Comparison of training dynamics on some examples between iFlood and AdaFlood.
    The ``Hard'' example is labeled \textit{horse}, but models usually predict \textit{cow};
    the ``Wrong'' example is incorrectly labeled in the dataset as \textit{cat} (there is no \textit{rat} class).
    }
    \label{fig:difficulty}
\end{figure*}

\paragraph{Motivation} 
Training samples are, however, not uniformly difficult: some are inherently easier to fit than others. 
\Cref{fig:dispersion_difficulty} shows the dispersion of difficulty on CIFAR10 and CIFAR100 with various levels of added label noise,
as measured by the heldout cross-entropy loss from cross-validated models.
Although difficulties on CIFAR10 without added noise are concentrated around difficulty $\leq 0.5$, as the noise increases, they vastly spread out.
CIFAR100 has a wide spread in difficulty, even without noise.
A constant flood level as used in iFlood may be reasonable for un-noised CIFAR10,
but it seems less appropriate for CIFAR100 or noisy-label cases.

Moreover, it may not be beneficial to aggressively drive the training loss for training samples that are outliers, noisy, or mislabeled. 
In \cref{fig:varying_difficulty}, we show training dynamics on an \textit{easy}, \textit{wrong}, and a \textit{hard} example from the training set of CIFAR10. 
With iFlood, each example's loss converges to the pre-determined flood level ($0.03$);
with AdaFlood, the \textit{easy} example converges towards zero loss,
while the \textit{wrong} and \textit{hard} examples maintain higher loss.

\subsection{Proposed Method: AdaFlood}
\label{subsec:adaflood}
\begin{figure}[t]
\begin{center}
   \includegraphics[width=0.95\textwidth]{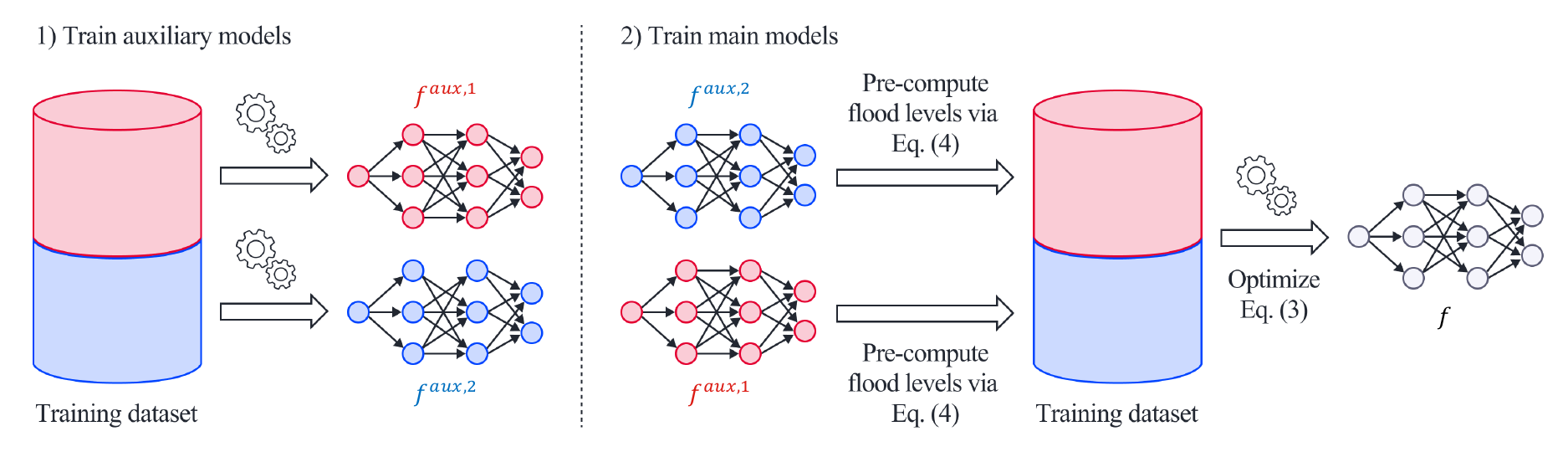}
\end{center}
\caption{AdaFlood with data-efficiency trick for settings where training data is limited and acquiring additional data is impractical. 
In the first stage, we partition the training set into two halves and train two auxiliary networks $f^{\text{aux},1}$ and $f^{\text{aux},2}$: one on each half. 
In the second stage, we use each auxiliary network to set the adaptive flood level of training samples from the half it has not seen, via Eq.~\ref{eq:aux}. 
The main network $f$ is then trained on the entire training set, minimizing the AdaFlood-regularized loss, Eq.~\ref{eq:l_adaflood}. 
Note that the flood levels are fixed over the course of training $f$ and need to be pre-computed once only. 
The cost of pre-computation can be further amortized over many training runs of $f$ with different configurations.}
\label{fig:training}
\end{figure}

Differences in per-sample difficulty are the basis of many advances in efficient neural network training and inference, such as batch or data selection \citep{coleman2020svp,fan2018learning,mindermann2022prioritized} and dynamic neural networks \citep{li2021dynamic,verelsttuytelaars2020}. 
AdaFlood connects this observation to flooding. 
Intuitively, easy training samples (e.g.\ a correctly-labeled image of a \textit{cat} in a typical pose) can be driven more aggressively to zero training loss without overfitting the model, while doing so for noisy, outlier, or incorrectly-labeled training samples may cause overfitting.
These types of data points behave differently during training \citep{zigzag},
and so should probably not be treated the same.
AdaFlood does so by setting a sample-specific flood level in its objective:
\begin{equation}
  \mathcal{L}_{\text{AdaFlood}} = \frac{1}{B}\sum_{i=1}^{B}\left(|\ell(y_i, f(\bm{x}_i)) - \theta_i| + \theta_i\right)
\label{eq:l_adaflood}
.\end{equation}
Here the sample-specific parameters $\theta_i$
should be set according to the individual sample's difficulty.
AdaFlood estimates this quantity according to
\begin{equation}
    \theta_i = \ell ( y_i, \phi_\gamma(f^{\mathrm{aux},i}(\bm{x}_i), y_i) )
\label{eq:aux}
,\end{equation}
where $f^{\mathrm{aux},i}$ is an auxiliary model trained with cross-validation
such that $\bm x_i$ is in its heldout set,
and $\phi_{\gamma}(\cdot)$ is a ``correction'' function explained in a moment.
\Cref{fig:training} illustrates the training process using \eqref{eq:l_adaflood},
\cref{app:held_out} gives further motivation,
and \cref{subsec:theory} gives further theoretical support.

The flood targets $\theta_i$ are fixed over the course of training the main network $f$, and can be pre-computed for each training sample prior to the first epoch of training $f$. 
We typically use five-fold cross-validation as a reasonable trade-off between computational expense
and good-enough models to estimate $\theta_i$, but see further discussion in \cref{subsec:efficient}.
The cost of this pre-processing step can be further amortized over many training runs of the main network $f$ since different variations and configurations of $f$ can reuse the adaptive flood levels.

\paragraph{Correction function.}
Unfortunately, the predictions from auxiliary models are not always correct even when trained on most of the training set -- if they were, our model would be perfect already.
In particular, the adaptive flood levels $\theta_i$ can be arbitrarily large for any difficult examples where the auxiliary model is incorrect;
this could lead to strange behavior when we encourage the primary model $f$ to be very incorrect.
We thus ``correct" the predictions with the correction function $\phi_\gamma$, which mixes between the dataset's label and the heldout model's signal.

For \textbf{regression tasks}, the predictions $f(\bm{x}_i) \in \R$ should simply be close to the labels $y_i \in \R$.
Here the correction function linearly interpolates the predictions and labels as,
\begin{align}
    \phi_\gamma(f^{\text{aux}}(\bm{x}_i), y_i) = (1-\gamma) f^{\text{aux}}(\bm{x}_i) + \gamma y_i
.\end{align}
Here $\gamma = 0$ fully trusts the auxiliary models (no ``correction''), while $\gamma = 1$ disables flooding.

For $K$-way \textbf{classification tasks}, $f(\bm{x}_i) \in \mathbb{R}^K$ is a vector of output probabilities (following a softmax layer), and the label is $y_i \in [K]$.
Cross-entropy loss only considers the probability of true class $y_i$: $\ell(y, \hat y) = -\log(\hat y_{y_i})$.
The $y_i$-th component of the correction function $\phi_\gamma(f^{\mathrm{aux}}(x_i), y_i)$ is then
\begin{align}
    \phi_\gamma(f^{\mathrm{aux},i}(\bm{x}_i), y_i)_{y_i} = (1-\gamma) f^{\mathrm{aux},i}(\bm{x}_i)_{y_i} + \gamma \, .
    \label{eq:phi-gamma-ce}
\end{align}
Again, for $\gamma = 0$ there is no ``correction,''
and for $\gamma = 1$ flooding is disabled,
as $\theta_i = - \log 1 = 0$.

The hyperparameter $\gamma \in [0, 1]$ is perhaps simpler to interpret and search for than directly identifying a flood level as in Flood or iFlood;
in those cases, the level is unbounded (in $[0, \infty)$ for supervised tasks and all of $\R$ for density estimation) and the choice is quite sensitive to the particular task.

\begin{algorithm}[t]
    \caption{Training of Auxiliary Network(s) and AdaFlood 
    }
    \begin{algorithmic}[1]
        \State Train a single auxiliary network $f^{\text{aux}}$ on the entire training set $\mathcal{D}$ \Comment Fine-tuning method only
        \For{ $\mathcal{D}^{\mathrm{aux},i}$ in $\{ \mathcal{D}^{\mathrm{aux},i} \}_{i=1}^n$ } 
        \State Train $f^{\text{aux}, i}$, either from scratch or by fine-tuning $f^{\text{aux}}$, on $\mathcal{D} \setminus \mathcal{D}^{\mathrm{aux},i}$
        \State Save the adaptive flood level $\theta_i$ for each $\bm x_i \in \mathcal D^{\mathrm{aux},i}$ using $f^{\text{aux}, i}$  on $\bm{x} \in \mathcal{D}^{\mathrm{aux},i}$ 
        \EndFor
        \State Train the main model $f$ using \cref{eq:l_adaflood} and adaptive flood levels $\theta$ computed above
    \end{algorithmic}
    \label{algo:adaflood} 
\end{algorithm}

\subsection{Efficiently Training Auxiliary Networks}
\label{subsec:efficient}

Although the losses from auxiliary networks can often be good measures for the difficulties of samples,
this is only true when the number of folds $n$ is reasonably large;
otherwise the training set of size about $\frac{n-1}{n} \lvert \mathcal D \rvert$
may be too much smaller than $\mathcal D$ for the model to have comparable performance.
The computational cost scales roughly linearly with $n$, however,
since we must train $n$ auxiliary networks:
if we do this in parallel it requires $n$ times the computational resources,
or if we do it sequentially it takes $n$ times as long as training a single model.
To alleviate the computational overhead for training auxiliary networks,
we sometimes instead approximate the process by fine-tuning a single auxiliary network.
More specifically, we first train a single base model $f^{\text{aux}}$ on the entire training set $\mathcal{D}$.
We then train each of the $n$ auxiliary models
by randomly re-initializing the last few layers,
then re-training with the relevant fold held out.
The process is illustrated in \cref{fig:fine_tuning} and \cref{algo:adaflood}.

Although this means that $\bm x_i$ does slightly influence the final prediction $f^{\mathrm{aux},i}(\bm x_i)$
(``training on the test set''),
it is worth remembering that we use $\theta_i$ only as a parameter in our model:
$\bm x_i$ is in fact a training data point for the overall model $f$ being trained.
This procedure is justified by recent understanding in the field
that in typical settings, a single data point only loosely influence the early layers of a network.
In highly over-parameterized settings (the ``kernel regime'') where neural tangent kernel theory is a good approximation to the training of $f^{\text{aux}}$ \citep{ntk},
re-initializing the last layer would completely remove the effect of $\bm x_i$ on the model.
Even in more realistic settings,
although the mechanism is not yet fully understood,
last layer re-training seems to do an excellent job at retaining ``core'' features and removing ``spurious'' ones that are more specific to individual data points \citep{kirichenko2023layer,labonte2023lastlayer}.

For smaller models with fewer than a million parameters,
we use $2$- or $5$-fold cross-validation,
since training multiple auxiliary models is not much of a computational burden.
For larger models such as ResNet18, however, we use the fine-tuning method.
This substantially reduces training time, since each fine-tuning gradient step is less expensive and the models converge much faster given strong features from lower levels than they do starting from scratch; \cref{subsec:ablation} gives a comparison.

To validate the quality of the flood levels from the fine-tuned auxiliary network, we compare them to the flood levels from $n=50$ auxiliary models using ResNet18~\citep{he2016deep} on CIFAR10~\citep{krizhevsky2009learning}; with $n = 50$, each model is being trained on 98\% of the full dataset, and thus should be a good approximation to the best that this kind of method can achieve.
The Spearman rank correlation between the fine-tuned method and the full cross-validation is $0.63$, a healthy indication that this method provides substantial signal for the ``correct'' $\theta_i$.
Our experimental results also reinforce that this procedure chooses a reasonable set of parameters.

\subsection{Theoretical Intuition}
\label{subsec:theory}

For a deeper understanding of AdaFlood's advantages,
we now examine a somewhat stylized supervised learning setting:
an overparameterized regime where the $\theta_i$ are nonetheless optimal.
\begin{prop}
Let $\mathcal F$ be a set of candidate models,
and suppose there exists an optimal model $f^* \in \argmin_{f \in \mathcal F} \E_{\bm x, y} \ell(y, f(\bm x))$,
where $\ell$ is a nonnegative loss function.
Given a dataset $\mathcal D = \{ (\bm x_i, y_i) \}_{i=1}^N$,
let $\hat f$ denote a minimizer of the empirical loss $\mathcal L(f) = \frac1N \sum_{i=1}^N \ell(y_i, f(\bm x_i))$;
suppose that, as in an overparameterized setting, $\mathcal L(\hat f) = 0$.
Also, let $\bar f$ be a minimizer of the AdaFlood loss \eqref{eq:l_adaflood}
using ``perfect'' flood levels $\theta_i = \ell(y_i, f^*(\bm x_i))$.
Then we have that
\begin{gather*}
    \mathcal L(\hat f) = 0 \le \mathcal L(f^*) = \mathcal L(\bar f)
    ,\quad
    \mathcal L_\text{AdaFlood}(\hat f) = 2 \mathcal L(f^*)
    \ge \mathcal L(f^*) = \mathcal L_\text{AdaFlood}(f^*) = \mathcal L_\text{AdaFlood}(\bar f)
.\end{gather*}
\end{prop}
We know that $\mathcal L(f^*)$ will be approximately the Bayes error, the irreducible distributional error achived by $f^*$;
this holds for instance by the law of large numbers, since $f^*$ is independent of the random sample $\mathcal D$.
Thus, if the Bayes error is nonzero and the $\theta_i$ are optimal,
we can see that empirical risk minimization of overparametrized models will find $\hat f$, and disallow $f^*$;
minimizing $\mathcal L_\text{AdaFlood}$, on the other hand, will allow the solution $f^*$ and disallow the empirical risk minimizer $\hat f$.

\begin{proof}
With this choice of $\theta_i$, we have that
\[
    \mathcal L_\text{AdaFlood}(f)
    = \frac1N \sum_{i=1}^N \Bigl( \lvert \ell(y_i, f(\bm x_i)) - \ell(y_i, f^*(\bm x_i)) \rvert + \ell(y_i f^*(\bm x_i)) \Bigr)
.\]
Since the absolute value is nonnegative, we have that
$\mathcal L_\text{AdaFlood}(f) \ge \mathcal L(f^*)$ for any $f$,
and that $\mathcal L_\text{AdaFlood}(f^*) = \mathcal L(f^*)$;
this establishes that $f^*$ minimizes $\mathcal L_\text{AdaFlood}$,
and that any minimizer $\bar f$ must achieve $\ell(y_i, \bar f(\bm x_i)) = \theta_i$ for each $i$,
so $\mathcal L(\bar f) = \mathcal L(f^*)$.
Using that $\ell(y_i, \hat f(\bm x_i)) = 0$ for each $i$,
as is necessary for $\ell \ge 0$ when $\mathcal L(\hat f) = 0$,
shows $\mathcal L_\text{AdaFlood}(\hat f) = \frac1N \sum_{i=1}^N 2 \theta_i = 2 \mathcal L(f^*)$.
\end{proof}

In settings where $\theta_i$ is not perfect (and we would not expect the auxiliary models to obtain \emph{perfect} estimates of the loss) the comparison will still approximately hold.
If $\theta_i$ consistently overestimates the $f^*$ loss, $f^*$ will still be preferred to $\hat f$: for instance, if $\theta_i = 2 \ell(y_i, f^*(\bm x_i))$, then $\mathcal L_\text{AdaFlood}(\hat f) = 4 \mathcal L(f^*) \ge 3 \mathcal L(f^*) = \mathcal L_\text{AdaFlood}(f^*)$.
On the other hand, if $\theta_i = \frac12 \ell(y_i, f^*(\bm x_i))$ -- a not-unreasonable situation when using a correction function -- then $\mathcal L_\text{AdaFlood}(\hat f) = \mathcal L(f^*) = \mathcal L_\text{AdaFlood}(f^*)$.
When $\theta_i$ is random, the situation is more complex,
but we can expect that noisy $\theta_i$ which somewhat overestimate the loss of $f^*$ will still prefer $f^*$ to $\hat f$.

\section{Experiments}
\label{sec:experiments}

We now demonstrate the effectiveness of AdaFlood on three tasks (probability density estimation, classification and regression) in four different domains (asynchronous event sequences, image, text and tabular).
We compare flooding methods on asynchronous event time in \cref{subsec:tpp} and image classification tasks in \cref{subsec:img_cls}.
We also demonstrate that AdaFlood is more robust to various noisy settings in \cref{subsec:noisy}, and that it yields better-calibrated models for image classification tasks in \cref{subsec:calibration}.
We investigate the performance of the fine-tuning scheme in \cref{subsec:ablation}.

\begin{table*}[t!]
    \centering
    \fontsize{9}{12.0}\selectfont
    \begin{tabu}{c|c|cc|ccc|ccc}
    \hline
    \multirow{2}{*}{NTPP} & 
    \multirow{2}{*}{Method} &
    \multicolumn{2}{c|}{Uber} &
    \multicolumn{3}{c|}{Reddit} &
    \multicolumn{3}{c}{Stack Overflow} \\
    & & RMSE & NLL  & RMSE  & NLL & ACC  & RMSE  & NLL & ACC \\
    \hline
    \multirow{8}{*}{Intensity-free} 
    & \multirow{2}{*}{Unrge.} & 75.83 & 3.86 & 0.25  & 1.28  & 55.26 & 6.69 & 3.66 & 45.52 \\ 
    & & (6.12) & (0.05) & (0.01) & (0.07) & (0.57) & (0.98) & (0.12) & (0.07)  \\
    & \multirow{2}{*}{Flood} & 64.34 & 4.01 & 0.25 & 1.17 & 57.46 &  4.12 & 3.46 & \bf{45.76} \\
    & & (3.85) & (0.02) & (0.01) & (0.06) & (0.84) & (0.23) & (0.03) & \bf{(0.03)} \\
    & \multirow{2}{*}{iFlood} & 67.07  & 3.97 & \bf{0.23}  & 1.11  & 56.59 & 4.12 & 3.46 & \bf{45.76} \\
    & & (3.12) & (0.06) & \bf{(0.01)} & (0.12) & (0.92) & (0.23) & (0.03) & \bf{(0.03)} \\
    & \multirow{2}{*}{AdaFlood} & \bf{59.69} & \bf{3.75} & 0.26 & \bf{1.09} & \bf{59.02} &  \bf{3.26} & \bf{3.45} & 45.67  \\
    & & \bf{(1.49)} & \bf{(0.01)} & (0.02) & \bf{(0.13)} & \bf{(0.91)} & \bf{(0.25)} & \bf{(0.04)} & (0.03) \\
    \hline
    \multirow{8}{*}{THP$^{+}$} 
    & \multirow{2}{*}{Unreg.} & 71.01 & 3.73 & 0.28 & 0.82 & 58.63 & 1.46 & 2.82  & 46.24 \\ 
    & & (6.12) & (0.05) & (0.01) & (0.07) & (0.57) & (0.98) & (0.12) & (0.07) \\
    & \multirow{2}{*}{Flood} & 68.61  & 3.70 & 0.26  & 1.02 & 58.05 & 1.39 & 2.79  & 46.31 \\
    & & (3.85) & (0.02) & (0.01) & (0.06) & (0.84) & (0.23) & (0.03) & (0.03)\\
    & \multirow{2}{*}{iFlood} & 68.61 & 3.70 & \bf{0.25} & 0.92 & 58.93 & 1.46  & 2.82 & 46.24 \\
    & & (4.76) & (0.17) & (0.01) & (0.23) & (1.26) & (0.06) & (0.04) & (0.08) \\
    & \multirow{2}{*}{AdaFlood} & \bf{54.85} & \bf{3.55} & \bf{0.25} & \bf{0.80} & \bf{61.34} & \bf{1.38} & \bf{2.77} & \bf{46.41} \\
    & & \bf{(1.49)} & \bf{(0.01)} & \bf{(0.02)} & \bf{(0.13)} & \bf{(0.91)} & \bf{(0.25)} & \bf{(0.04)} & \bf{(0.03)} \\
    \Xhline{2\arrayrulewidth}  
    \end{tabu}
    \caption{Comparison of flooding methods on asynchronous event sequence datasets. The numbers are the means and standard errors (in parentheses) over three runs.}
    \label{tbl:tpp}
\end{table*}

\subsection{Results on Asynchronous Event Sequences}
\label{subsec:tpp}

In this section, we compare flooding methods on asynchronous event sequence datasets of which goal is to estimate the probability distribution of the next event time given the previous event times.
Each event may or may not have a class label.
Asynchronous event sequences are often modeled as temporal point processes and terms are used interchangeably. 
Details are provided in \cref{app:datasets}.

\paragraph{Datasets}
We use two popular benchmark datasets, Stack Overflow (predicting the times at which users receive badges) and Reddit (predicting posting times).
Following~\cite{bae2023meta}, we also benchmark our method on a dataset with stronger periodic patterns: Uber (predicting pick-up times).
We split each training dataset into train $(80\%)$ and validation $(20\%)$ sets.

Following the literature in temporal point processes (TPPs), we use two metrics to evaluate TPP models: \textit{root mean squared error} (RMSE) and \textit{negative log-likelihood} (NLL). 
While NLL can be misleadingly low if the probability density is mostly focused on the correct event time, RMSE is not a good metric if stochastic components of TPPs are ignored and a baseline is directly trained on the ground truth event times. Therefore, we train our TPP models on NLL and use RMSE at test time to ensure that we do not rely too heavily on RMSE scores and account for the stochastic nature of TPPs.
When class labels for events are available, we also report the accuracy of class predictions.

\paragraph{Implementation}
For TPP models to predict the asynchronous event times, we employ Intensity-free models~\citep{shchur2019intensity} based on GRU~\citep{gru2014chung}, and Transformer Hawkes Processes (THP)~\citep{zuo2020transformer} based on Transformer~\citep{attention2017vaswani}. 
THP predicts intensities to compute log-likelihood and expected event times, but this approach can be computationally expensive due to the need to compute integrals, particularly double integrals to calculate the expected event times. 
To overcome this challenge while maintaining performance, we follow \citet{bae2023meta} in using a mixture of log-normal distributions, proposed in \cite{shchur2019intensity}, for the decoder; we call this THP$^+$. 
The optimal flood levels are selected via a grid search on $\{-50, -45, -40 \dots, 0, 5 \} \cup \{-4, -3 \dots, 3, 4\} $ for Flood and iFlood, and optimal $\gamma$ on $\{0.0, 0.1 \dots, 0.9 \} $ for AdaFlood using the validation set.
We use five auxiliary models.

\paragraph{Results}
In order to evaluate the effectiveness of various regularization methods, we present the results of our experiments in \cref{tbl:tpp}
(showing means and standard errors from three runs). 
This is the first time we know of where flooding methods have been applied in this domain;
we see that all flooding methods improve the generalization performance here, sometimes substantially.
Further, AdaFlood significantly outperforms the other methods for most models on most datasets, suggesting that the instance-wise flooding level adaptation using auxiliary models is a particularly effective way to enhance the generalization capabilities of both TPP models.

\begin{table*}[t!]
    \centering
    \fontsize{8.7}{11.7}\selectfont
    \begin{tabu}{c|c|c|c|c|c|c}
    \hline
    \multirow{2}{*}{Method} &
    \multicolumn{2}{c|}{SVHN} &
    \multicolumn{2}{c|}{CIFAR10} &
    \multicolumn{2}{c}{CIFAR100} \\
    \cline{2-7}
    & w/o $L_2$ reg. & w/ $L_2$ reg. & w/o $L_2$ reg. & w/ $L_2$ reg. & w/o $L_2$ reg. & w/ $L_2$ reg. \\
    \hline
    \hline
    Unreg. & 95.65 $\pm$ 0.05 & 96.07 $\pm$ 0.01 & 87.80 $\pm$ 0.31 & 90.35 $\pm$ 0.21 & 56.59 $\pm$ 0.32 & 61.49 $\pm$ 0.16 \\ 
    Flood & 95.63 $\pm$ 0.02 & 96.13 $\pm$ 0.02 & 87.57 $\pm$ 0.16 & 90.09 $\pm$ 0.20 & 55.88 $\pm$ 0.18 & 60.96 $\pm$ 0.03 \\
    iFlood & 95.63 $\pm$ 0.08 & 96.05 $\pm$ 0.02 & 87.96 $\pm$ 0.07 & 90.57 $\pm$ 0.12 & 56.32 $\pm$ 0.05 & 61.63 $\pm$ 0.12 \\
    AdaFlood & \bf{95.72 $\pm$ 0.01} & \bf{96.16 $\pm$ 0.02} & \bf{88.38 $\pm$ 0.18} & \bf{90.82 $\pm$ 0.08} & \bf{57.25 $\pm$ 0.14} & \bf{62.31 $\pm$ 0.14} \\
    \Xhline{2\arrayrulewidth}
    \end{tabu}
    \caption{Comparison of flooding methods on image classification datasets with and without $L_2$ regularization. The numbers are the means and standard errors over three runs.}
    \label{tbl:cls}
\end{table*}

\subsection{Results on Image Classification}
\label{subsec:img_cls}

\paragraph{Datasets}
We use SVHN~\citep{svhn2011netzer}, CIFAR-10, and CIFAR 100~\citep{krizhevsky2009learning} as the benchmarks for image classification with random crop and horizontal flip as augmentation.
Unlike \citet{iflood2022xie2022}, we split each training dataset into train ($80\%$) and validation ($20\%$) sets for hyperparameter search; thus our numbers are generally somewhat worse than what they reported.

\paragraph{Implementation}
On the image classification datasets, following \citet{flood2020ishida} and similar to \citet{iflood2022xie2022}, we consider training ResNet18~\citep{he2016deep} on the datasets with and without $L_2$ regularization (with a weight of $10^{-4}$).
All methods are trained with SGD for $300$ epochs, with early stopping.
We use a multi-step learning rate scheduler with an initial learning rate of 0.1 and decay coefficient of 0.2, applied at every 60 epochs.
The optimal flood levels are selected based on validation performace with a grid search on $\{0.01, 0.02 \dots, 0.1, 0.15, 0.2 \dots, 1.0 \} $ for Flood and iFlood, and $\{0.05, 0.1 \dots, 0.95 \} $ for AdaFlood.
We use a single ResNet18 auxiliary network where its layer 3 and 4 are randomly initialized and fine-tuned on held-out sets with $n=10$ splits.

\paragraph{Results}
The results are presented in \cref{tbl:cls}.
We report the means and standard errors of accuracies over three runs. 
We can observe that flooding methods, including AdaFlood, are not significantly better than the unregularized baseline on SVHN.
However, AdaFlood noticeably improves the performance over the other methods on harder datasets like CIFAR10 and CIFAR100, whereas iFlood is not obviously better than the baseline and Flood is worse than the baseline on CIFAR100.

\subsection{Noisy Labels}
\label{subsec:noisy}

\paragraph{Datasets}
In addition to CIFAR10 for image classification, we also use the tabular datasets Brazilian Houses and Wine Quality from OpenML~\citep{OpenML2013}, following \citet{tree2022grinsztajn}, for regression tasks.
We further employ Stanford Sentiment Treebank (SST-2) for the text classification task, following \citet{iflood2022xie2022}.
Details of datasets are provided in \cref{app:datasets}. 

We report accuracy for classification tasks.
For regression tasks, we report \textit{mean squared error} (MSE) in the main body, as well as \textit{mean absolute error} (MAE) and \textit{R}$^2$ score in \cref{fig:tabular_diff_metrics}.

\paragraph{Implementation}
We inject noise for both image and text classification by changing the label to a uniformly randomly selected wrong class, following \citet{iflood2022xie2022}. 
More specifically, for $\alpha \%$ of the training data, we change the label to a uniformly random class other than the original label.
For the regression tasks, we add errors sampled from a skewed normal distribution, with skewness parameter ranging from $0.0$ to $3.0$.

\begin{figure*}[t!]
    \begin{subfigure}[t]{0.245\textwidth}
        \centering
        \includegraphics[width=0.99\textwidth]{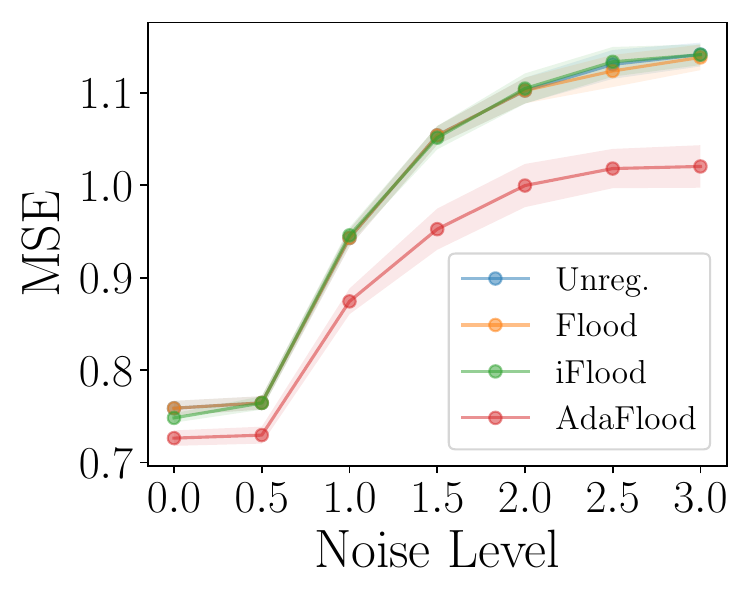}
        \caption{Wine Quality}
        \label{fig:wine_noisy}
    \end{subfigure}
    \begin{subfigure}[t]{0.245\textwidth}
        \centering
        \includegraphics[width=0.99\textwidth]{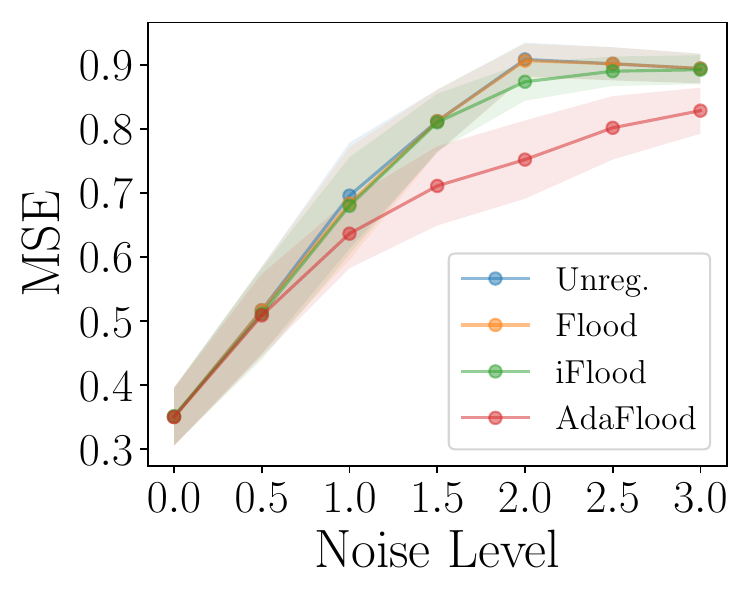}
        \caption{Brazilian Houses}
        \label{fig:house_noisy}
    \end{subfigure}
    \begin{subfigure}[t]{0.245\textwidth}
        \centering
        \includegraphics[width=0.99\textwidth]{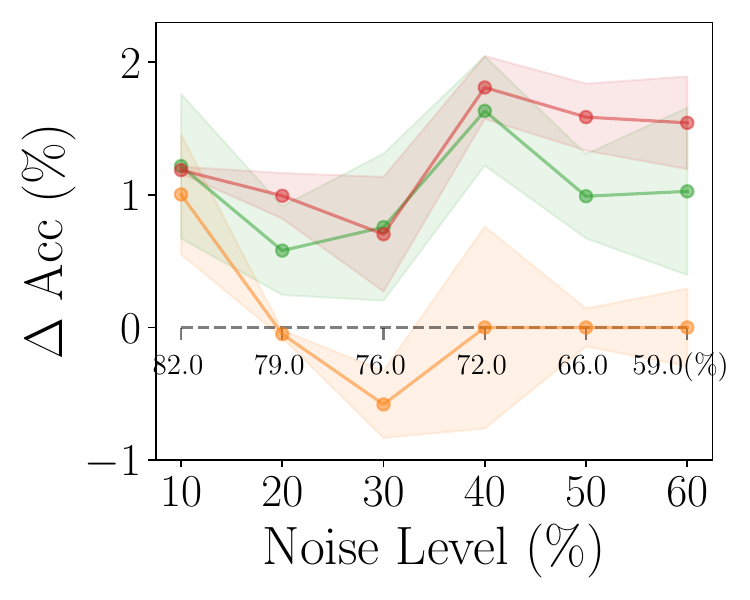}
        \caption{CIFAR10}
        \label{fig:cifar10_noisy}
    \end{subfigure}
    \begin{subfigure}[t]{0.245\textwidth}
        \centering
        \includegraphics[width=0.99\textwidth]{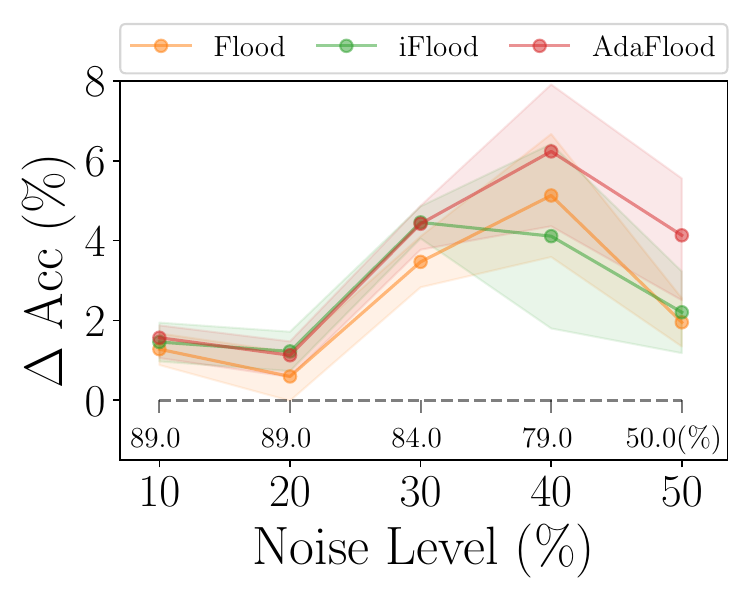}
        \caption{SST-2}
        \label{fig:text_noisy}
    \end{subfigure}
    \caption{Comparison of flooding methods on tabular and image datasets with noise and bias.}
    \label{fig:noisy}
\end{figure*}

\paragraph{Results}
\cref{fig:noisy} compares the flooding methods for noisy settings. 
We report the mean and standard error over three runs for CIFAR10, and five and seven runs for tabular datasets and SST-2, respectively.
We provide $\Delta Acc$ $(\%)$ for CIFAR10 and SST-2 compared to the unregularized model: that is, we plot the accuracy of each method minus the accuracy of the unregularized method, to display the gaps between methods more clearly.
The mean accuracies of the unregularized method are displayed below the zero line.

\begin{itemize}
    \item Wine Quality, \cref{fig:wine_noisy}: AdaFlood slightly outperforms the other methods at first, but the gap significantly increases as the noise level increases.
    \item Brazilian Houses, \cref{fig:house_noisy}: There is no significant difference between the methods for small noise level, \eg noise parameter $\leq 1.5$, but MSE for AdaFlood becomes significantly lower as the noise level increases.
    \item CIFAR10, \cref{fig:cifar10_noisy}: iFlood and AdaFlood significantly outperform Flood and unregularized. 
    AdaFlood also outperforms iFlood when the noise level is high (\eg $\geq 50 \%$).
    \item SST-2, \cref{fig:text_noisy}: Flooding methods significantly outperform the unregularized approach.
    AdaFlood is comparable to iFlood up to the noise level of $30\%$, but noticeably outperforms it as the noise level further increases.
\end{itemize}

Overall, AdaFlood is more robust to noise than the other flooding methods, since the model pays less attention to those samples with wrong or noisy labels. %

\begin{figure*}[t!]
    \begin{subfigure}[t]{0.245\textwidth}
        \centering
        \includegraphics[width=0.99\textwidth]{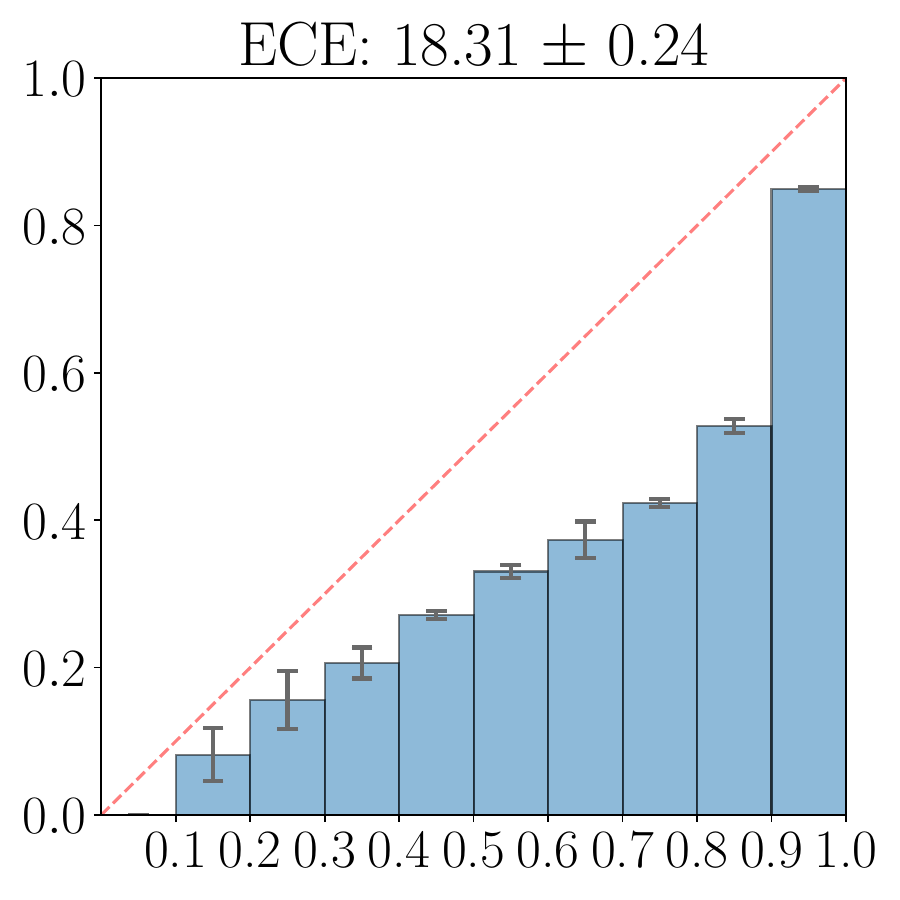}
        \caption{Unregularized}
        \label{fig:cali_unreg}
    \end{subfigure}
    \begin{subfigure}[t]{0.245\textwidth}
        \centering
        \includegraphics[width=0.99\textwidth]{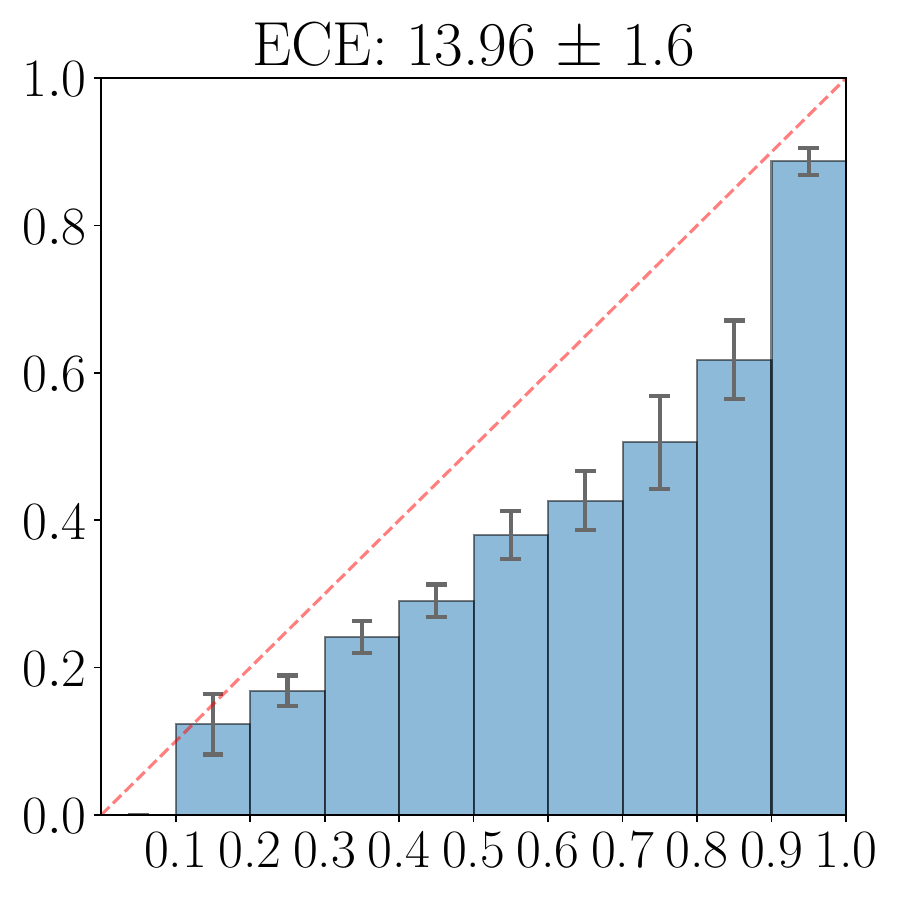}
        \caption{Flood}
        \label{fig:cali_flood}
    \end{subfigure}
    \begin{subfigure}[t]{0.245\textwidth}
        \centering
        \includegraphics[width=0.99\textwidth]{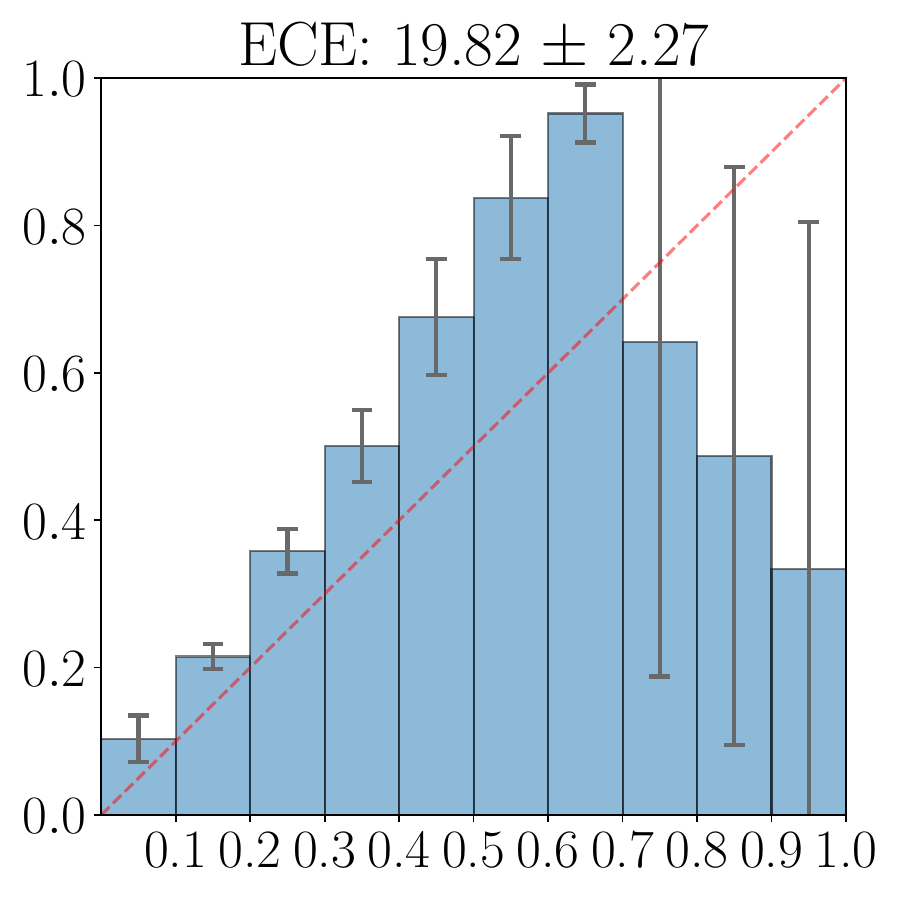}
        \caption{iFlood}
        \label{fig:cali_iflood}
    \end{subfigure}
    \begin{subfigure}[t]{0.245\textwidth}
        \centering
        \includegraphics[width=0.99\textwidth]{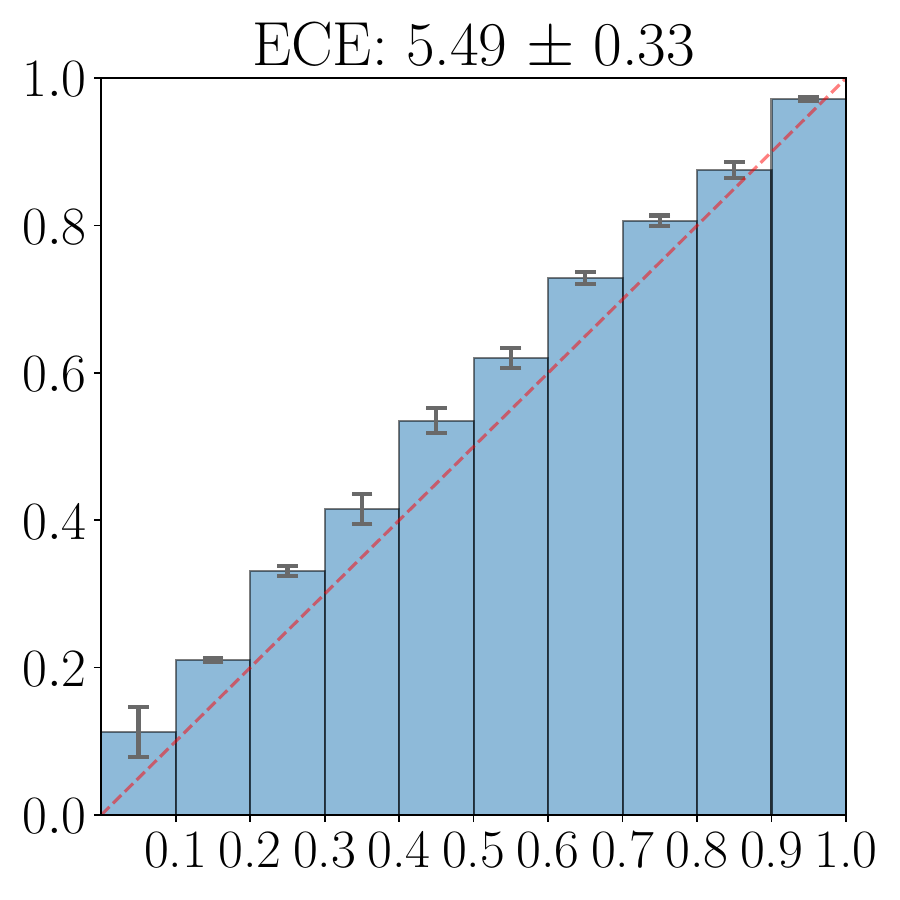}
        \caption{AdaFlood}
        \label{fig:cali_adaflood}
    \end{subfigure}
    \caption{Calibration results of flooding methods with $10$ bins on CIFAR100. The bars and errors are the means and standard errors over three runs, respectively.}
    \label{fig:calibration_cifar100}
\end{figure*}

\subsection{Calibration}
\label{subsec:calibration}

\paragraph{Datasets and implementation}
Miscalibration -- neural networks being over or under-confident -- has been a well-known issue in deep learning.
We thus evaluate the quality of calibration with different flooding methods on CIFAR100,
as measured by the Expected Calibration Error (ECE) metric.
(\Cref{fig:calibration_cifar10} does the same for CIFAR10,
but since model predictions are usually quite confident, this becomes difficult to measure.)

We use a ResNet18 with $L_2$ regularization with the optimal hyperparameters for the baseline and flooding methods.
The optimal hyperparameter varies by seed for each run.

\paragraph{Result}
\cref{fig:calibration_cifar100} provides the calibration quality in ECE metric as well as a visualization over three runs, compared to perfect calibration (dotted red lines).
We can observe that AdaFlood significantly improves the calibration, both in ECE and visually.
Note that iFlood significantly miscalibrates at the bins corresponding to high probability \eg bin $\geq 0.7$, compared to the other methods, and also has high standard errors.
This behavior is expected, since iFlood encourages the model not to predict higher than a probability of $\exp(-b)$, where $b$ denotes the flood level used in iFlood.

\subsection{Ablation study: Fine-tuning vs. Multiple Auxiliaries}
\label{subsec:ablation}

\begin{wrapfigure}{r}{0.45\textwidth}
  \vspace{-7mm}
  \begin{center}
    \includegraphics[width=0.45\textwidth,valign=t]{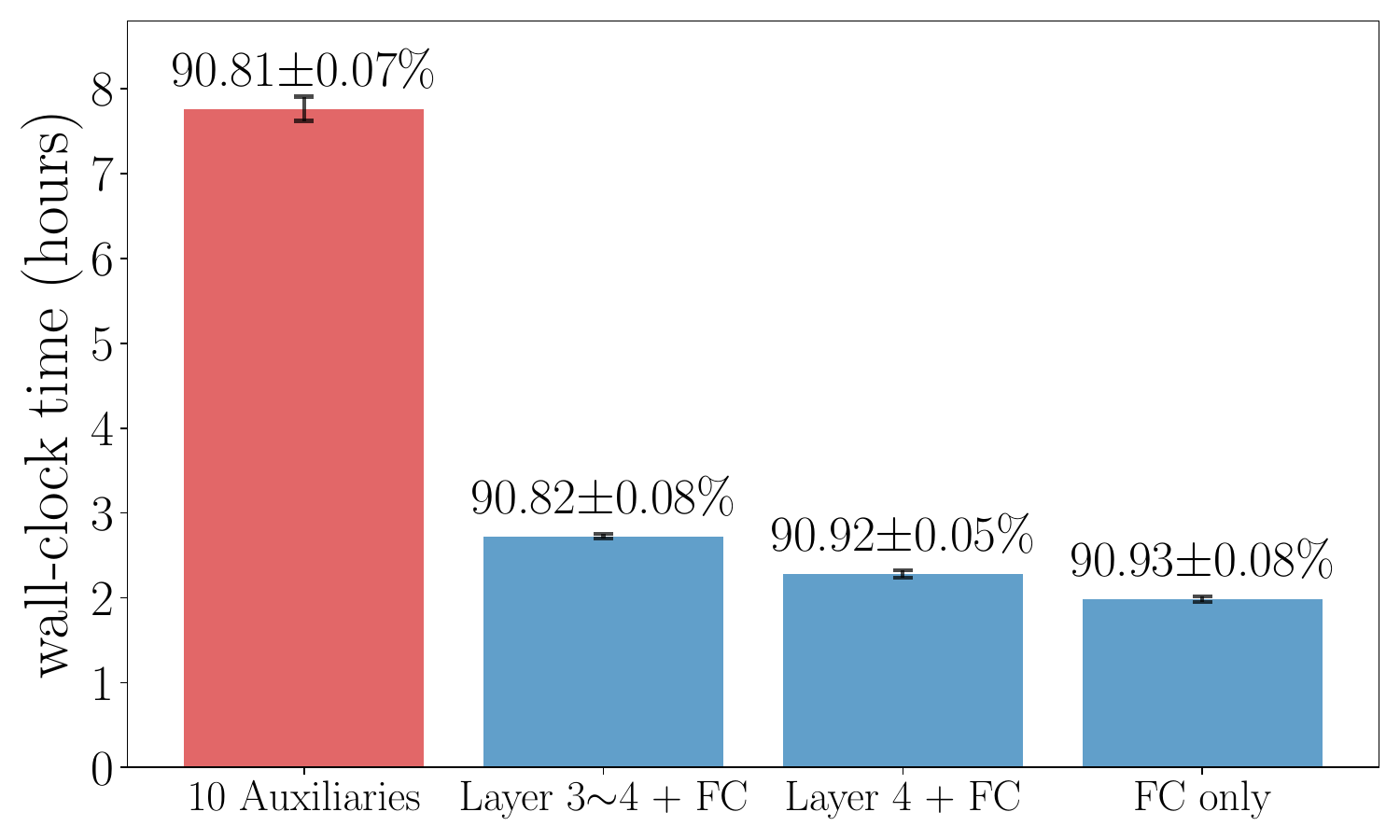}
  \end{center}
  \vspace{-5mm}
  \caption{Comparison of aux. training}
  \label{fig:ablation_finetune}
  \vspace{-3mm}
\end{wrapfigure}
\cref{fig:ablation_finetune} compares training of ten ResNet18 auxiliary networks (original proposal) to the single fine-tuned auxiliary network (efficient variant) in terms of wall-clock time for training using an NVIDIA GTX 1080 Ti GPU, and performance of the corresponding main model, on CIFAR10.
Furthermore, we provide the training time and performance of an auxiliary network fine-tuned on different layers: $Layer3, 4 + FC$, $Layer3 + FC$, and $FC$, where $Layer3$ and $4$ are the $3$rd, $4$th layers in ResNet18 and $FC$ denotes the last fully connected layer.

The result shows that training multiple auxiliary networks yields the same-quality model as fine-tuning, although training time is about $3 - 4$ times longer.
There is also little difference in performance between different fine-tuning methods:
it seems that fine-tuning only the FC layer is sufficient to forget the samples,
with early-stopping regularizing well enough for similar generalization ability.

\section{Conclusion}
\label{sec:conclusion}

In this paper, we introduced the Adaptive Flooding (AdaFlood) regularizer, a novel reguralization technique that adaptively regularizes a loss for each sample based on the difficulty of the sample. 
Each flood level is computed only once through an auxiliary training procedure with held-out splitting, which we can make more efficient by fine-tuning the last few layers on held-out sets. 
Experimental results on various domains and tasks: density estimation for asynchronous event sequences, image and text classification tasks as well as regression tasks on tabular datasets, with and without noise, demonstrated that our approach is more robustly applicable to a varied range of tasks including calibration.

\paragraph{Reproducibility}
For each experiment, we listed implementation details of the experiment such as model, optimizer, learning rate scheduler, regularization, and search space for hyperparameters. 
We also specify datasets we used for each experiment, and how they were split and augmented, along with the description of metrics.
The code will be released in the final version.

\ifdefstring{\docstatus}{submission}{}{
\subsubsection*{Acknowledgements}
This work was supported in part by
the Natural Sciences and Engineering Resource Council of Canada,
the Canada CIFAR AI Chairs program,
the BC DRI Group,
Compute Ontario,
and the Digital Resource Alliance of Canada.
}

\bibliography{iclr2024_conference}

\begin{thebibliography}{52}
\providecommand{\natexlab}[1]{#1}
\providecommand{\url}[1]{\texttt{#1}}
\expandafter\ifx\csname urlstyle\endcsname\relax
  \providecommand{\doi}[1]{doi: #1}\else
  \providecommand{\doi}{doi: \begingroup \urlstyle{rm}\Url}\fi

\bibitem[Bae et~al.(2023)Bae, Ahmed, Tung, and Oliveira]{bae2023meta}
Wonho Bae, Mohamed~Osama Ahmed, Frederick Tung, and Gabriel~L Oliveira.
\newblock Meta temporal point processes.
\newblock In \emph{ICLR}, 2023.

\bibitem[Balestriero et~al.(2022)Balestriero, Bottou, and
  LeCun]{balestriero2022effects}
Randall Balestriero, Leon Bottou, and Yann LeCun.
\newblock The effects of regularization and data augmentation are class
  dependent.
\newblock In \emph{NeurIPS}, 2022.

\bibitem[Belkin et~al.(2019)Belkin, Hsu, Ma, and Mandal]{belkin2019reconciling}
Mikhail Belkin, Daniel Hsu, Siyuan Ma, and Soumik Mandal.
\newblock Reconciling modern machine-learning practice and the classical
  bias--variance trade-off.
\newblock \emph{Proceedings of the National Academy of Sciences}, 116\penalty0
  (32):\penalty0 15849--15854, 2019.

\bibitem[Bertinetto et~al.(2919)Bertinetto, Henriques, Torr, and
  Vedaldi]{bertinetto2019meta}
Luca Bertinetto, Jo{\~a}o Henriques, Philip H.~S. Torr, and Andrea Vedaldi.
\newblock Meta-learning with differentiable closed-form solvers.
\newblock In \emph{ICLR}, 2919.

\bibitem[Cao et~al.(2019)Cao, Wei, Gaidon, Arechiga, and Ma]{cao2019imbalance}
Kaidi Cao, Colin Wei, Adrien Gaidon, Nikos Arechiga, and Tengyu Ma.
\newblock Learning imbalanced datasets with label-distribution-aware margin
  loss.
\newblock In \emph{NeurIPS}, 2019.

\bibitem[Chung et~al.(2014)Chung, Gulcehre, Cho, and Bengio]{gru2014chung}
Junyoung Chung, Caglar Gulcehre, KyungHyun Cho, and Yoshua Bengio.
\newblock Empirical evaluation of gated recurrent neural networks on sequence
  modeling.
\newblock In \emph{NeurIPS}, 2014.

\bibitem[Coleman et~al.(2020)Coleman, Yeh, Mussmann, Mirzasoleiman, Bailis,
  Liang, Leskovec, and Zaharia]{coleman2020svp}
Cody Coleman, Christopher Yeh, Stephen Mussmann, Baharan Mirzasoleiman, Peter
  Bailis, Percy Liang, Jure Leskovec, and Matei Zaharia.
\newblock Selection via proxy: Efficient data selection for deep learning.
\newblock In \emph{ICLR}, 2020.

\bibitem[Ding et~al.(2019)Ding, Chin, Liu, and
  Marculescu]{ding2019regularizing}
Ruizhou Ding, Ting-Wu Chin, Zeye Liu, and Diana Marculescu.
\newblock Regularizing activation distribution for training binarized deep
  networks.
\newblock In \emph{CVPR}, 2019.

\bibitem[Du et~al.(2016)Du, Dai, Trivedi, Upadhyay, Gomez-Rodriguez, and
  Song]{du2016recurrent}
Nan Du, Hanjun Dai, Rakshit Trivedi, Utkarsh Upadhyay, Manuel Gomez-Rodriguez,
  and Le~Song.
\newblock Recurrent marked temporal point processes: Embedding event history to
  vector.
\newblock In \emph{KDD}, 2016.

\bibitem[Fan et~al.(2018)Fan, Tian, Qin, Li, and Liu]{fan2018learning}
Yang Fan, Fei Tian, Tao Qin, Xiang-Yang Li, and Tie-Yan Liu.
\newblock Learning to teach.
\newblock In \emph{ICLR}, 2018.

\bibitem[Franceschi et~al.(2018)Franceschi, Frasconi, Salvo, Grazzi, and
  Pontil]{franceschi2018bilevel}
Luca Franceschi, Paolo Frasconi, Saverio Salvo, Riccardo Grazzi, and
  Massimiliano Pontil.
\newblock Bilevel programming for hyperparameter optimization and
  meta-learning.
\newblock In \emph{ICML}, 2018.

\bibitem[Gong et~al.(2022)Gong, Mori, and Tung]{gong2022ranksim}
Yu~Gong, Greg Mori, and Frederick Tung.
\newblock {RankSim}: Ranking similarity regularization for deep imbalanced
  regression.
\newblock In \emph{ICML}, 2022.

\bibitem[Grinsztajn et~al.(2022)Grinsztajn, Oyallon, and
  Varoquaux]{tree2022grinsztajn}
L{\'e}o Grinsztajn, Edouard Oyallon, and Ga{\"e}l Varoquaux.
\newblock Why do tree-based models still outperform deep learning on typical
  tabular data?
\newblock In \emph{NeurIPS}, 2022.

\bibitem[Hanson \& Pratt(1988)Hanson and Pratt]{hanson1988comparing}
Stephen Hanson and Lorien Pratt.
\newblock Comparing biases for minimal network construction with
  back-propagation.
\newblock In \emph{NeurIPS}, 1988.

\bibitem[He et~al.(2016)He, Zhang, Ren, and Sun]{he2016deep}
Kaiming He, Xiangyu Zhang, Shaoqing Ren, and Jian Sun.
\newblock Deep residual learning for image recognition.
\newblock In \emph{CVPR}, 2016.

\bibitem[Ioffe \& Szegedy(2015)Ioffe and Szegedy]{ioffe2015batch}
Sergey Ioffe and Christian Szegedy.
\newblock Batch normalization: Accelerating deep network training by reducing
  internal covariate shift.
\newblock In \emph{ICML}, 2015.

\bibitem[Ishida et~al.(2020)Ishida, Yamane, Sakai, Niu, and
  Sugiyama]{flood2020ishida}
Takashi Ishida, Ikko Yamane, Tomoya Sakai, Gang Niu, and Masashi Sugiyama.
\newblock Do we need zero training loss after achieving zero training error?
\newblock In \emph{ICML}, 2020.

\bibitem[Jacot et~al.(2018)Jacot, Gabriel, and Hongler]{ntk}
Arthur Jacot, Franck Gabriel, and Cl{\'{e}}ment Hongler.
\newblock Neural tangent kernel: Convergence and generalization in neural
  networks.
\newblock In \emph{NeurIPS}, 2018.

\bibitem[Jiang et~al.(2021)Jiang, Zhang, Talwar, and Mozer]{cscore}
Ziheng Jiang, Chiyuan Zhang, Kunal Talwar, and Michael~C Mozer.
\newblock Characterizing structural regularities of labeled data in
  overparameterized models.
\newblock In \emph{ICML}, 2021.

\bibitem[Kirichenko et~al.(2023)Kirichenko, Izmailov, and
  Wilson]{kirichenko2023layer}
Polina Kirichenko, Pavel Izmailov, and Andrew~Gordon Wilson.
\newblock Last layer re-training is sufficient for robustness to spurious
  correlations.
\newblock In \emph{ICLR}, 2023.

\bibitem[Krizhevsky et~al.(2009)Krizhevsky, Hinton,
  et~al.]{krizhevsky2009learning}
Alex Krizhevsky, Geoffrey Hinton, et~al.
\newblock Learning multiple layers of features from tiny images.
\newblock 2009.

\bibitem[Krogh \& Hertz(1991)Krogh and Hertz]{krogh1991weight}
Anders Krogh and John~A. Hertz.
\newblock A simple weight decay can improve generalization.
\newblock In \emph{NeurIPS}, 1991.

\bibitem[LaBonte et~al.(2023)LaBonte, Muthukumar, and
  Kumar]{labonte2023lastlayer}
Tyler LaBonte, Vidya Muthukumar, and Abhishek Kumar.
\newblock Towards last-layer retraining for group robustness with fewer
  annotations.
\newblock In \emph{NeurIPS}, 2023.

\bibitem[Li et~al.(2021)Li, Wang, Wang, Liang, Li, and Chang]{li2021dynamic}
Changlin Li, Guangrun Wang, Bing Wang, Xiaodan Liang, Zhihui Li, and Xiaojun
  Chang.
\newblock Dynamic slimmable network.
\newblock In \emph{CVPR}, 2021.

\bibitem[Li et~al.(2020)Li, Gu, Mayer, Gool, and Timofte]{li2020group}
Yawei Li, Shuhang Gu, Christoph Mayer, Luc~Van Gool, and Radu Timofte.
\newblock Group sparsity: The hinge between filter pruning and decomposition
  for network compression.
\newblock In \emph{CVPR}, 2020.

\bibitem[Liang et~al.(2021)Liang, Wu, Li, Wang, Meng, Qin, Chen, Zhang, and
  Liu]{liang2021rdrop}
Xiaobo Liang, Lijun Wu, Juntao Li, Yue Wang, Qi~Meng, Tao Qin, Wei Chen, Min
  Zhang, and Tie-Yan Liu.
\newblock {R-Drop}: Regularized dropout for neural networks.
\newblock In \emph{NeurIPS}, 2021.

\bibitem[Lim et~al.(2022)Lim, Erichson, Utrera, Xu, and Mahoney]{lim2022noisy}
Soon~Hoe Lim, N.~Benjamin Erichson, Francisco Utrera, Winnie Xu, and Michael~W.
  Mahoney.
\newblock Noisy feature mixup.
\newblock In \emph{ICLR}, 2022.

\bibitem[Liu et~al.(2019)Liu, Simonyan, and Yang]{liu2019darts}
Hanxiao Liu, Karen Simonyan, and Yiming Yang.
\newblock {DARTS}: Differentiable architecture search.
\newblock In \emph{ICLR}, 2019.

\bibitem[Liu \& Ye(2010)Liu and Ye]{lq2010liu2010}
Jun Liu and Jieping Ye.
\newblock Efficient l1/lq norm regularization.
\newblock \emph{arXiv preprint arXiv:1009.4766}, 2010.

\bibitem[Maini et~al.(2022)Maini, Garg, Lipton, and
  Kolter]{maini2022characterizing}
Pratyush Maini, Saurabh Garg, Zachary Lipton, and J~Zico Kolter.
\newblock Characterizing datapoints via second-split forgetting.
\newblock \emph{Advances in Neural Information Processing Systems},
  35:\penalty0 30044--30057, 2022.

\bibitem[Mindermann et~al.(2022)Mindermann, Brauner, Razzak, Sharma, Kirsch,
  Xu, H{\"o}ltgen, Gomez, Morisot, Farquhar, and
  Gal]{mindermann2022prioritized}
S{\"o}ren Mindermann, Jan~M Brauner, Muhammed~T Razzak, Mrinank Sharma, Andreas
  Kirsch, Winnie Xu, Benedikt H{\"o}ltgen, Aidan~N Gomez, Adrien Morisot,
  Sebastian Farquhar, and Yarin Gal.
\newblock Prioritized training on points that are learnable, worth learning,
  and not yet learnt.
\newblock In \emph{ICML}, 2022.

\bibitem[Nakkiran et~al.(2021)Nakkiran, Kaplun, Bansal, Yang, Barak, and
  Sutskever]{nakkiran2021deep}
Preetum Nakkiran, Gal Kaplun, Yamini Bansal, Tristan Yang, Boaz Barak, and Ilya
  Sutskever.
\newblock Deep double descent: Where bigger models and more data hurt.
\newblock \emph{Journal of Statistical Mechanics: Theory and Experiment},
  2021\penalty0 (12):\penalty0 124003, 2021.

\bibitem[Netzer et~al.(2011)Netzer, Wang, Coates, Bissacco, Wu, and
  Ng]{svhn2011netzer}
Yuval Netzer, Tao Wang, Adam Coates, Alessandro Bissacco, Bo~Wu, and Andrew~Y
  Ng.
\newblock Reading digits in natural images with unsupervised feature learning.
\newblock In \emph{NIPS Workshop on Deep Learning and Unsupervised Feature
  Learning}, 2011.

\bibitem[Neyshabur et~al.(2015)Neyshabur, Tomioka, and
  Srebro]{neyshabur2015search}
Behnam Neyshabur, Ryota Tomioka, and Nathan Srebro.
\newblock In search of the real inductive bias: On the role of implicit
  regularization in deep learning.
\newblock In \emph{ICLR Workshop}, 2015.

\bibitem[Ren et~al.(2022)Ren, Guo, and Sutherland]{zigzag}
Yi~Ren, Shangmin Guo, and Danica~J. Sutherland.
\newblock Better supervisory signals by observing learning paths.
\newblock In \emph{ICLR}, 2022.

\bibitem[Shchur et~al.(2020)Shchur, Bilo{\v{s}}, and
  G{\"u}nnemann]{shchur2019intensity}
Oleksandr Shchur, Marin Bilo{\v{s}}, and Stephan G{\"u}nnemann.
\newblock Intensity-free learning of temporal point processes.
\newblock In \emph{ICLR}, 2020.

\bibitem[Srivastava et~al.(2014)Srivastava, Hinton, Krizhevsky, Sutskever, and
  Salakhutdinov]{srivastava2014dropout}
Nitish Srivastava, Geoffrey Hinton, Alex Krizhevsky, Ilya Sutskever, and Ruslan
  Salakhutdinov.
\newblock Dropout: a simple way to prevent neural networks from overfitting.
\newblock \emph{JMLR}, pp.\  1929--1958, 2014.

\bibitem[Szegedy et~al.(2016)Szegedy, Vanhoucke, Ioffe, Shlens, and
  Wojna]{szegedy2016rethinking}
Christian Szegedy, Vincent Vanhoucke, Sergey Ioffe, Jon Shlens, and Zbigniew
  Wojna.
\newblock Rethinking the {I}nception architecture for computer vision.
\newblock In \emph{CVPR}, 2016.

\bibitem[Tibshirani(1996)]{tibshirani1996regression}
Robert Tibshirani.
\newblock Regression shrinkage and selection via the {L}asso.
\newblock \emph{Journal of the Royal Statistical Society: Series B
  (Methodological)}, 58\penalty0 (1):\penalty0 267--288, 1996.

\bibitem[Van~der Maaten \& Hinton(2008)Van~der Maaten and Hinton]{tsne}
Laurens Van~der Maaten and Geoffrey Hinton.
\newblock Visualizing data using {t-SNE}.
\newblock \emph{Journal of machine learning research}, 9\penalty0 (11), 2008.

\bibitem[Vanschoren et~al.(2013)Vanschoren, van Rijn, Bischl, and
  Torgo]{OpenML2013}
Joaquin Vanschoren, Jan~N. van Rijn, Bernd Bischl, and Luis Torgo.
\newblock Openml: networked science in machine learning.
\newblock \emph{SIGKDD Explorations}, pp.\  49--60, 2013.

\bibitem[Vaswani et~al.(2017)Vaswani, Shazeer, Parmar, Uszkoreit, Jones, Gomez,
  Kaiser, and Polosukhin]{attention2017vaswani}
Ashish Vaswani, Noam Shazeer, Niki Parmar, Jakob Uszkoreit, Llion Jones,
  Aidan~N Gomez, {\L}ukasz Kaiser, and Illia Polosukhin.
\newblock Attention is all you need.
\newblock In \emph{NeurIPS}, 2017.

\bibitem[Verelst \& Tuytelaars(2020)Verelst and
  Tuytelaars]{verelsttuytelaars2020}
Thomas Verelst and Tinne Tuytelaars.
\newblock Dynamic convolutions: Exploiting spatial sparsity for faster
  inference.
\newblock In \emph{CVPR}, 2020.

\bibitem[Verma et~al.(2019)Verma, Lamb, Beckham, Najafi, Mitliagkas, Lopez-Paz,
  and Bengio]{verma2019manifold}
Vikas Verma, Alex Lamb, Christopher Beckham, Amir Najafi, Ioannis Mitliagkas,
  David Lopez-Paz, and Yoshua Bengio.
\newblock Manifold mixup: Better representations by interpolating hidden
  states.
\newblock In \emph{ICML}, 2019.

\bibitem[Wager et~al.(2013)Wager, Wang, and Liang]{wager2013dropout}
Stefan Wager, Sida Wang, and Percy~S Liang.
\newblock Dropout training as adaptive regularization.
\newblock In \emph{NeurIPS}, 2013.

\bibitem[Wang et~al.(2021)Wang, Cheng, Chen, Tang, and
  Hsieh]{wang2021rethinking}
Ruochen Wang, Minhao Cheng, Xiangning Chen, Xiaocheng Tang, and Cho-Jui Hsieh.
\newblock Rethinking architecture selection in differentiable {NAS}.
\newblock In \emph{ICLR}, 2021.

\bibitem[Xie et~al.(2022)Xie, Zhen, Li, Zhang, Zhou, and
  Ding]{iflood2022xie2022}
Yuexiang Xie, WANG Zhen, Yaliang Li, Ce~Zhang, Jingren Zhou, and Bolin Ding.
\newblock {iFlood}: A stable and effective regularizer.
\newblock In \emph{ICLR}, 2022.

\bibitem[Yuan et~al.(2020)Yuan, Tay, Li, Wang, and Feng]{yuan2020revisiting}
Li~Yuan, Francis E.~H. Tay, Guilin Li, Tao Wang, and Jiashi Feng.
\newblock Revisiting knowledge distillation via label smoothing regularization.
\newblock In \emph{CVPR}, 2020.

\bibitem[Zhang et~al.(2021)Zhang, Bengio, Hardt, Recht, and
  Vinyals]{zhang2021understanding}
Chiyuan Zhang, Samy Bengio, Moritz Hardt, Benjamin Recht, and Oriol Vinyals.
\newblock Understanding deep learning (still) requires rethinking
  generalization.
\newblock \emph{Communications of the ACM}, 64\penalty0 (3):\penalty0 107--115,
  2021.

\bibitem[Zhang et~al.(2018)Zhang, Cisse, Dauphin, and
  Lopez-Paz]{zhang2017mixup}
Hongyi Zhang, Moustapha Cisse, Yann~N Dauphin, and David Lopez-Paz.
\newblock mixup: Beyond empirical risk minimization.
\newblock In \emph{ICLR}, 2018.

\bibitem[Zhuang et~al.(2020)Zhuang, Zhang, Huang, Zeng, Shuang, and
  Li]{zhuang2020neuron}
Tao Zhuang, Zhixuan Zhang, Yuheng Huang, Xiaoyi Zeng, Kai Shuang, and Xiang Li.
\newblock Neuron-level structured pruning using polarization regularizer.
\newblock In \emph{NeurIPS}, 2020.

\bibitem[Zuo et~al.(2020)Zuo, Jiang, Li, Zhao, and Zha]{zuo2020transformer}
Simiao Zuo, Haoming Jiang, Zichong Li, Tuo Zhao, and Hongyuan Zha.
\newblock Transformer {H}awkes process.
\newblock In \emph{ICML}, 2020.

\end{thebibliography}
\bibliographystyle{iclr2024_conference}

\clearpage
\appendix

\section{Why we calculate $\theta$ using held-out data}
\label{app:held_out}

In \cref{subsec:adaflood},
we estimate $\theta_i$ for each training sample using the output of an auxiliary network $f^\text{aux}(x_i)$ that is trained on a held-out dataset.
In fact, this adaptive flood level $\theta_i$ can be considered as the sample difficulty when training the main network.
Hence, it is reasonable to consider existing difficulty measurements based on learning dynamics, like C-score \citep{cscore} or forgetting score \citep{maini2022characterizing}.
However, we find these methods are not robust when wrong labels exist in the training data,
because the network will learn to remember the wrong label of $\bm x_i$,
and hence provide a low $\theta_i$ for the wrong sample, which is harmful to our method.
That is why we propose to split the whole training set into $n$ parts and train $f^\text{aux}(\bm x_i)$ for $n$ times (each with different $n-1$ parts).

\begin{figure}[h]
    \centering
    \includegraphics[width=1\columnwidth,trim=80 0 70 0, clip]{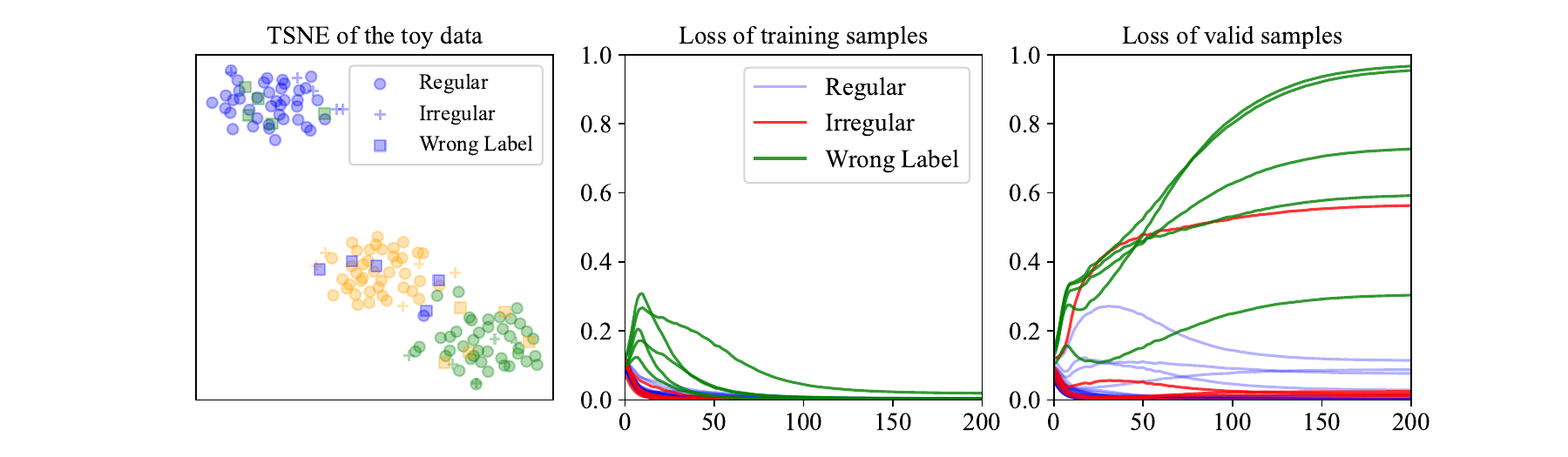}
    \caption{Left: the t-SNE \citep{tsne} of toy Gaussian example; middle: loss of different samples in the training set; right: loss of different samples in the validation set.}
    \label{fig:diff_loss}
\end{figure}

\paragraph{Dataset and implementation}
To verify this,
we conduct experiments on a toy Gaussian dataset,
as illustrated in the first panel in \cref{fig:diff_loss}.
Assume we have $N$ samples, each sample in 2-tuple ($x, y$). 
To draw a sample, we first select the label $y=k$ following a uniform distribution over all $K$ classes. After that, we sample the input signal $x \mid (y=k)\sim\mathcal{N}({\mu}_k,\sigma^2I)$, where $\sigma$ is the noise level for all the samples. ${\mu}_k$ is the mean vector for all the samples in class $k$. Each ${\mu}_k$ is a 10-dim vector, in which each dimension is randomly selected from $\{-\delta_\mu,0,\delta_\mu\}$. Such a process is similar to selecting 10 different features for each class.
We consider 3 types of samples for each class:
regular samples, the typical or easy samples in our training set, have a small $\sigma$;
irregular samples have a larger $\sigma$;
mislabeled samples have a small $\sigma$, but with a flipped label.
We generate two datasets following this same procedure (call them datasets $A$ and $B$).
The, we randomly initialize a 2-layer MLP with ReLU layers and train it on dataset $A$.
At the end of every epoch, we record the loss of each sample in dataset $A$.

\paragraph{Result}
The learning paths are illustrated in the second panel in \cref{fig:diff_loss}.
The model is clearly able remember all the wrong labels,
as all the curves converge to a small value.
If we calculate $\theta_i$ in this way,
all $\theta_i$ would have similar values.
However,
if we instead train the model using dataset $B$,
which comes from the same distribution but is different from dataset $A$,
the learning curves of samples in dataset $A$ will behave like the last panel in \cref{fig:diff_loss}.
The mislabeled and some irregular samples can be clearly identified from the figure.
Calculating $\theta_i$ in this way gives different samples more distinct flood values,
which makes our method more robust to sample noise,
as our experiments on various scenarios show.

\begin{figure}[t]
\begin{center}
   \includegraphics[width=0.70\textwidth]{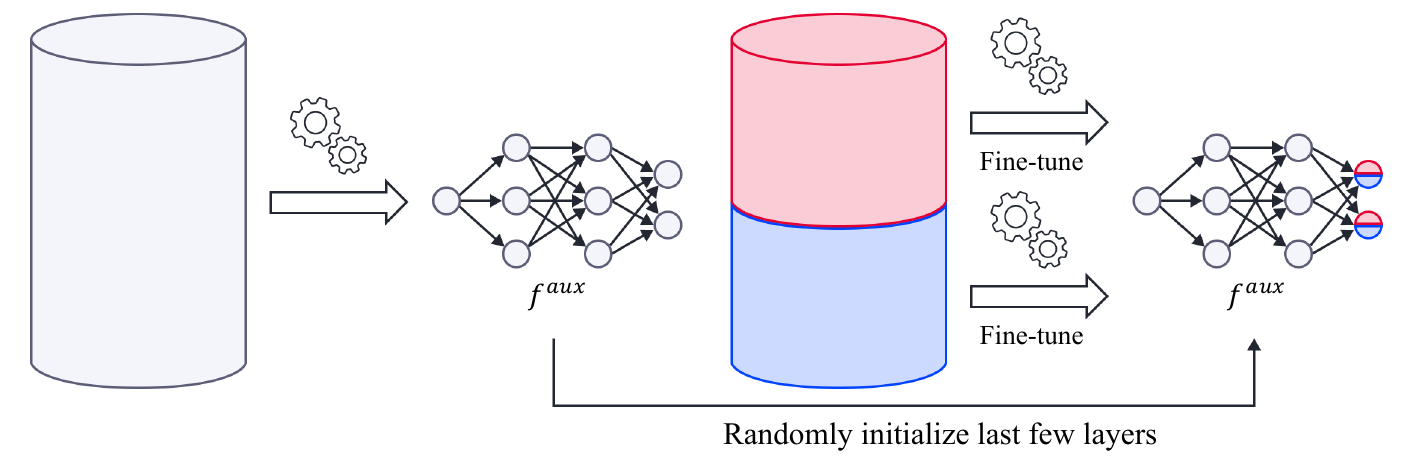}
\end{center}
\vspace{-3mm}
\caption{Efficient fine-tuning method for training a auxiliary network when held-out split is $n=2$.}
\label{fig:fine_tuning}
\end{figure}

\begin{figure*}[t!]
    \begin{subfigure}[t]{0.245\textwidth}
        \centering
        \includegraphics[width=0.99\textwidth]{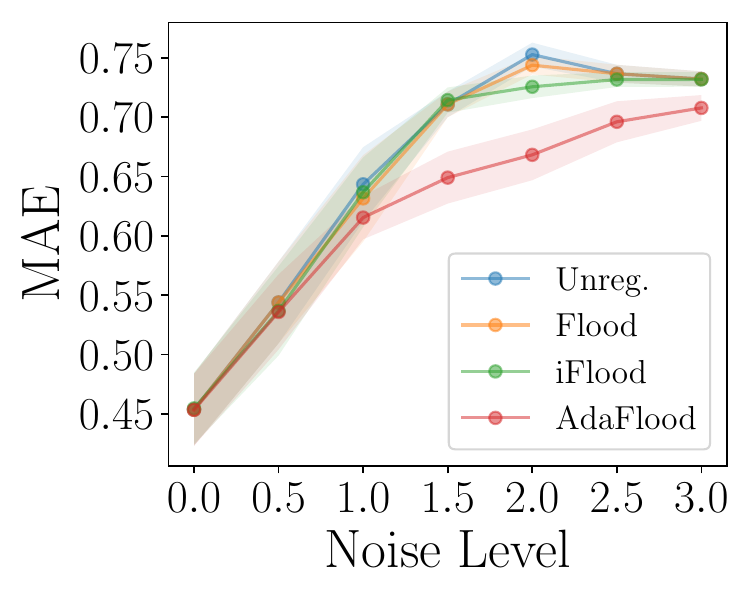}
        \caption{Brazilian House (MAE)}
        \label{fig:house_noisy_mae}
    \end{subfigure}
    \hfill
    \begin{subfigure}[t]{0.245\textwidth}
        \centering
        \includegraphics[width=0.99\textwidth]{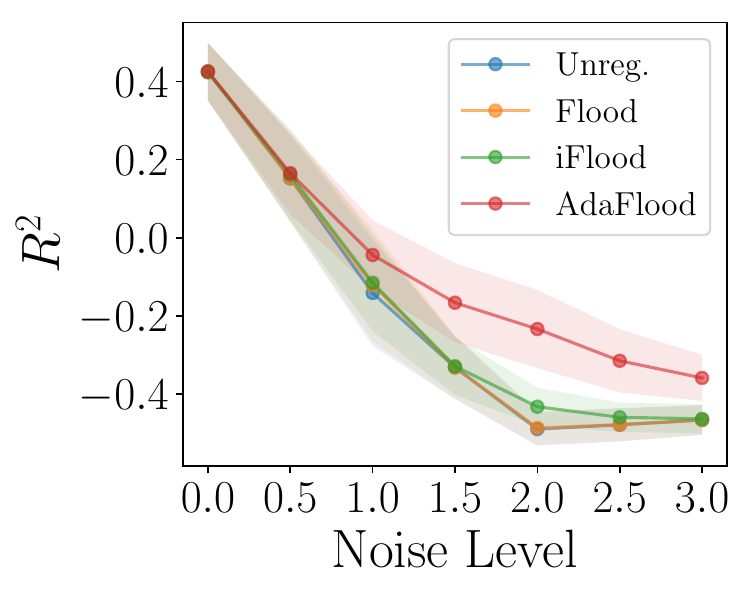}
        \caption{Brazilian House (R$^2$)}
        \label{fig:house_noisy_r2}
    \end{subfigure}
    \begin{subfigure}[t]{0.245\textwidth}
        \centering
        \includegraphics[width=0.99\textwidth]{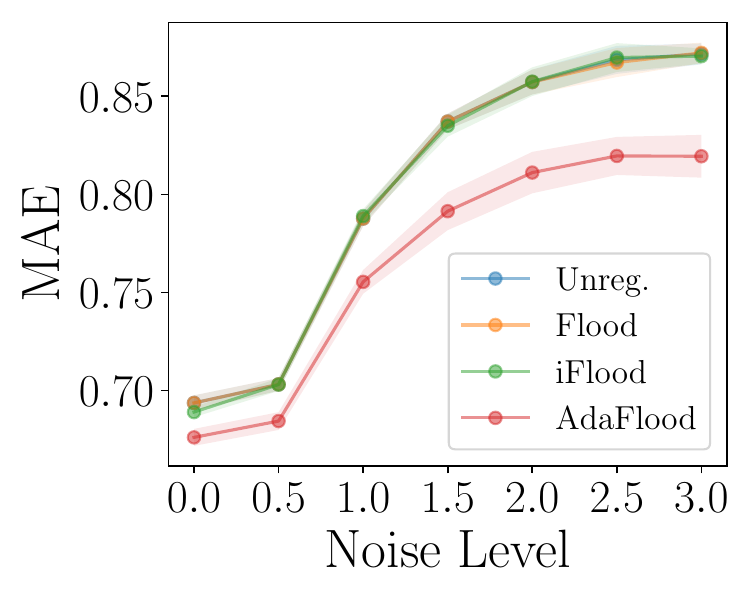}
        \caption{Wine Quality (MAE)}
        \label{fig:wine_noisy_mae}
    \end{subfigure}
    \hfill
    \begin{subfigure}[t]{0.245\textwidth}
        \centering
        \includegraphics[width=0.99\textwidth]{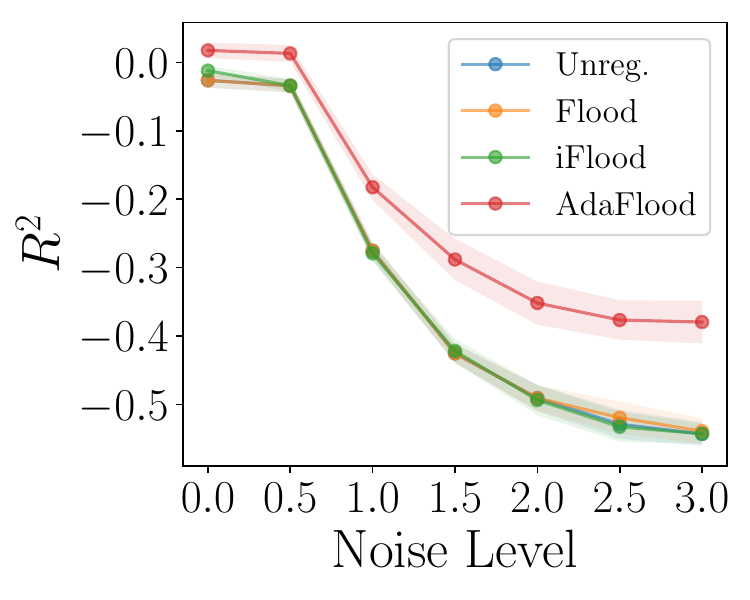}
        \caption{Wine Quality (R$^2$)}
        \label{fig:wine_noisy_r2}
    \end{subfigure}
    \caption{Additional results in various metrics on tabular datasets with noise and bias}
    \label{fig:tabular_diff_metrics}
\end{figure*}

\begin{figure*}[t!]
    \begin{subfigure}[t]{0.245\textwidth}
        \centering
        \includegraphics[width=0.99\textwidth]{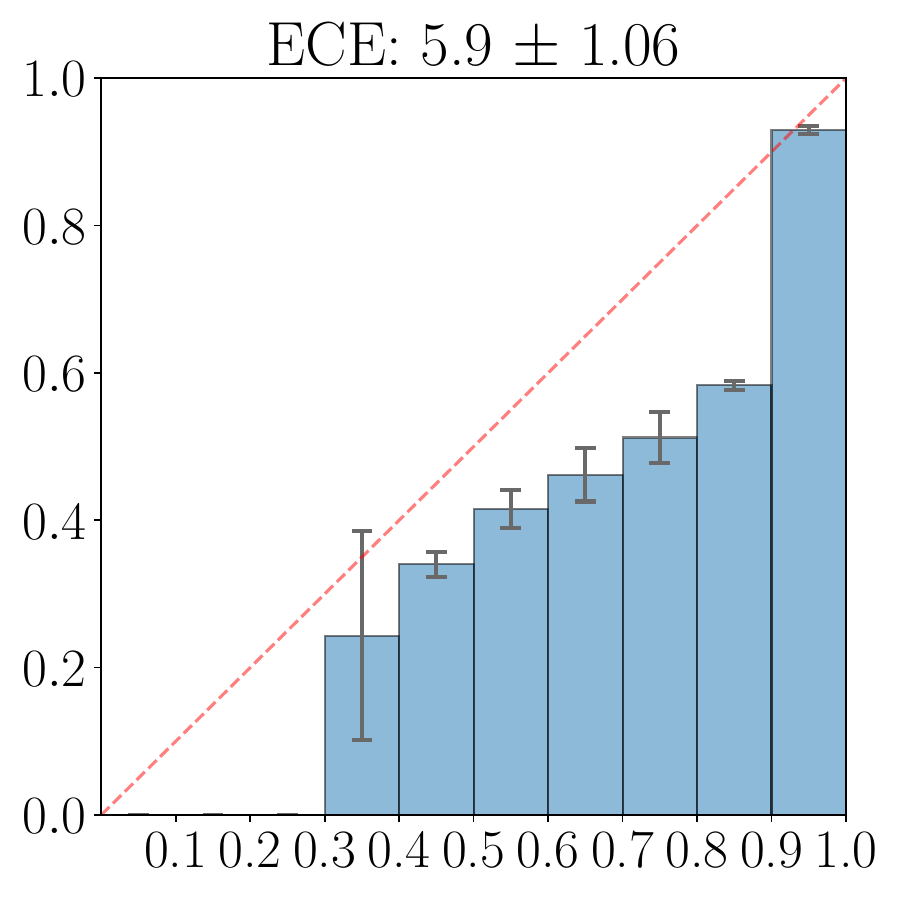}
        \caption{Unregularized}
        \label{app:fig:cali_unreg}
    \end{subfigure}
    \begin{subfigure}[t]{0.245\textwidth}
        \centering
        \includegraphics[width=0.99\textwidth]{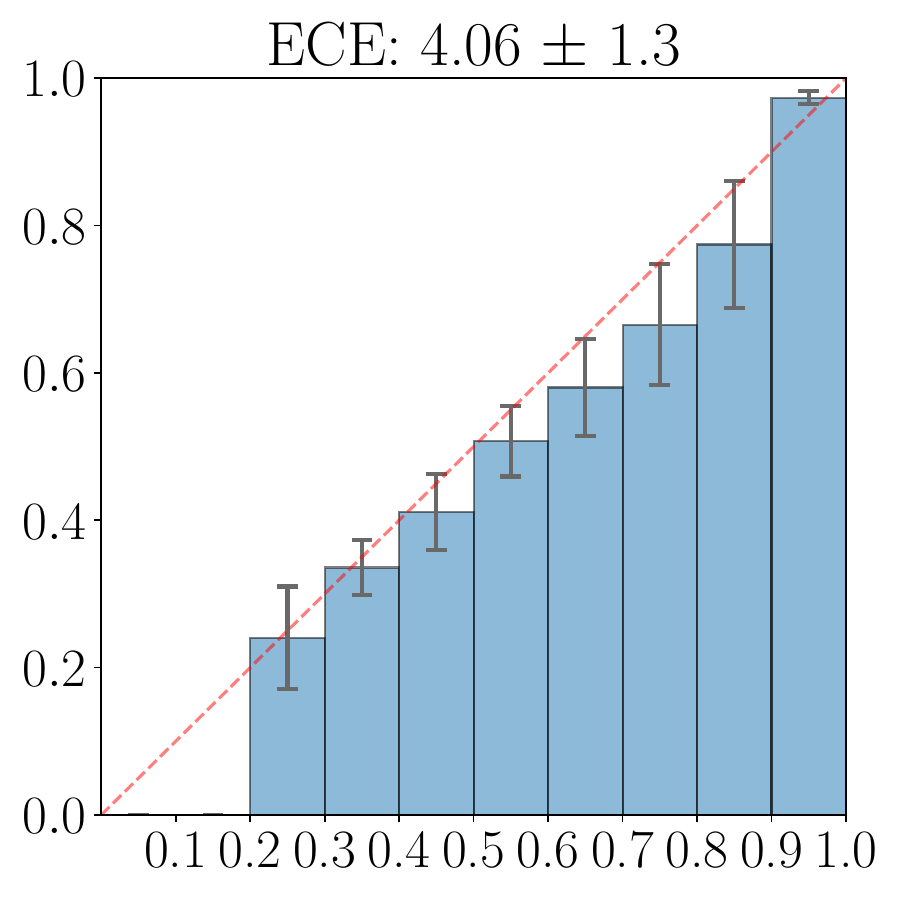}
        \caption{Flood}
        \label{app:fig:cali_flood}
    \end{subfigure}
    \begin{subfigure}[t]{0.245\textwidth}
        \centering
        \includegraphics[width=0.99\textwidth]{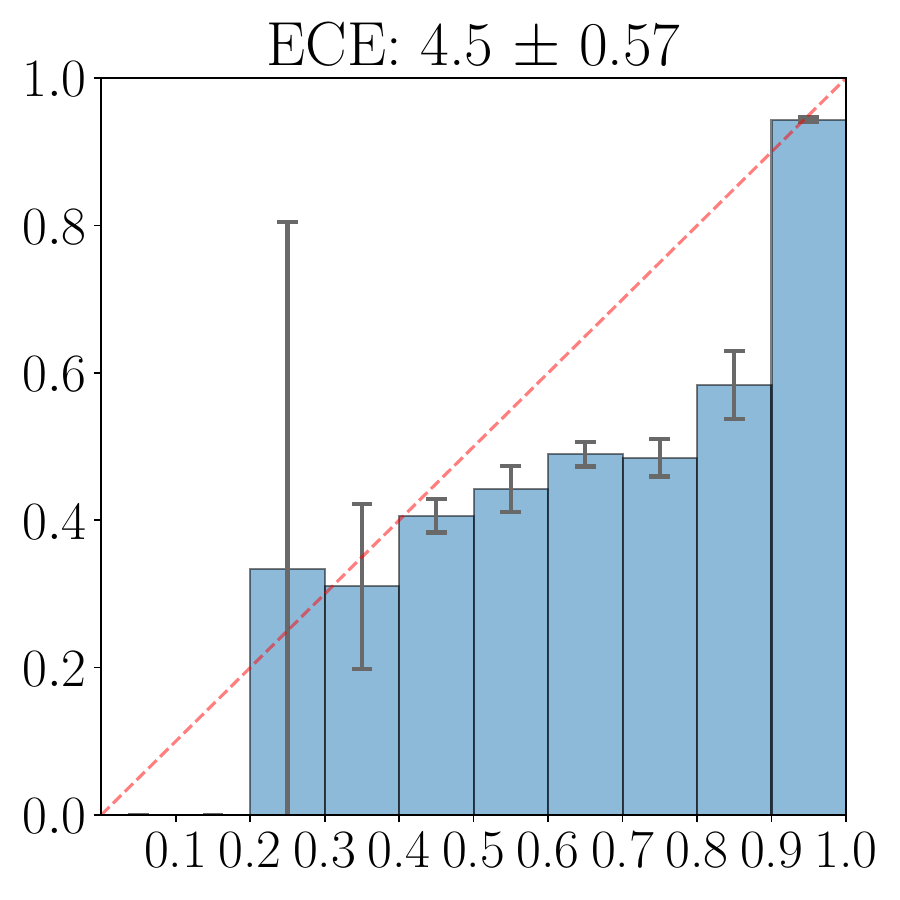}
        \caption{iFlood}
        \label{app:fig:cali_iflood}
    \end{subfigure}
    \begin{subfigure}[t]{0.245\textwidth}
        \centering
        \includegraphics[width=0.99\textwidth]{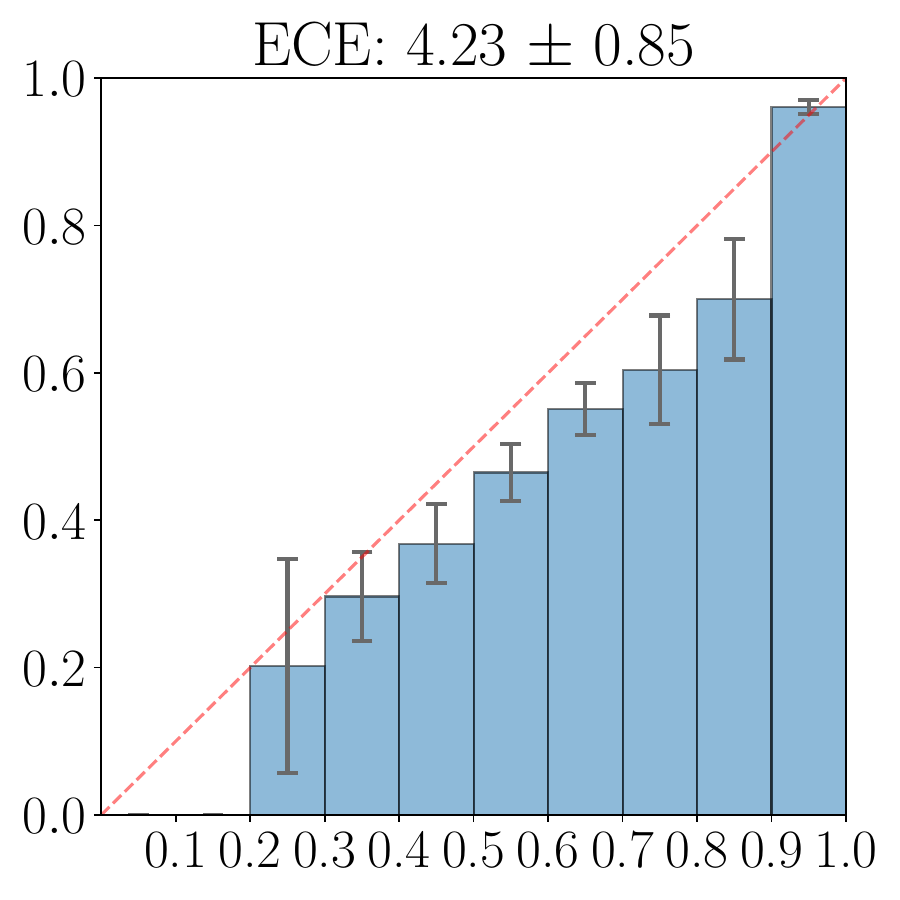}
        \caption{AdaFlood}
        \label{app:fig:cali_adaflood}
    \end{subfigure}
    \caption{Calibration results of flooding methods with $10$ bins on CIFAR10.}
    \label{fig:calibration_cifar10}
\end{figure*}

\section{Details about Datasets}
\label{app:datasets}

\paragraph{Stack Overflow}
It contains $6{,}633$ sequences with $480{,}414$ events where an event is the acquisition of badges received by users.
The maximum number of sequence length is $736$ and the number of marks is $22$.
The dataset is provided by \citet{du2016recurrent}; we use the first folder, following \citet{shchur2019intensity} and \citet{bae2023meta}.

\paragraph{Reddit}
It contains $10{,}000$ sequences with $532{,}026$ events where an event is posting in Reddit.
The maximum number of sequence length is $736$ and the number of marks is $22$.
Marks represent sub-reddit categories.

\paragraph{Uber}
It contains $791$ sequences with $701{,}579$ events where an event is pick-up of customers.
The maximum number of sequence length is $2{,}977$ and there is no marks.
It is processed and provided by \citet{bae2023meta}.

\paragraph{Brazilian Houses}
It contains information of $10{,}962$ houses to rent in Brazil in 2020 with $13$ features.
The target is the rent price for each house in Brazilian Real.
According to OpenML~\citep{OpenML2013} where we obtained this dataset, since the data is web-scrapped, there are some values in the dataset that can be considered outliers.

\paragraph{Wine Quality}
It contains $6{,}497$ samples with $11$ features and the quality of wine is numerically labeled as targets.
This dataset is also obtained from OpenML~\citep{OpenML2013}.

\paragraph{SST-2}
The Stanford Sentiment Treebank (SST-2) is a dataset containing fully annotated parse trees, enabling a comprehensive exploration of how sentiment influences language composition. 
Comprising $11{,}855$ individual sentences extracted from film reviews, this dataset underwent parsing using the Stanford parser, resulting in a collection of $215{,}154$ distinct phrases. 

\end{document}